%% file: main.tex
\pdfoutput=1

\documentclass[11pt]{article}

\usepackage[final]{acl}

\usepackage{times}
\usepackage{latexsym}

\usepackage{inconsolata}

\usepackage{algorithm}
\usepackage{algorithmic}
\usepackage{amsmath}
\usepackage{booktabs}
\usepackage{multirow}
\usepackage{subcaption}
\usepackage{enumitem}
\usepackage{wrapfig}
\usepackage{mathtools}
\usepackage{makecell}
\usepackage{hyperref}
\usepackage{xcolor}
\usepackage[normalem]{ulem}
\usepackage{rotating}
\usepackage{tablefootnote}
\usepackage{float}
\usepackage{color-edits}
\usepackage{comment}
\usepackage{soul}
\usepackage{titling}
\usepackage{amssymb}%

\newcommand{\modelname}{SEA}

\usepackage[T1]{fontenc}

\usepackage[utf8]{inputenc}

\usepackage{microtype}

\usepackage{inconsolata}

\usepackage{graphicx}

\definecolor{veronica-red}{RGB}{196,30,58}

\title{
\includegraphics[width=0.8cm]{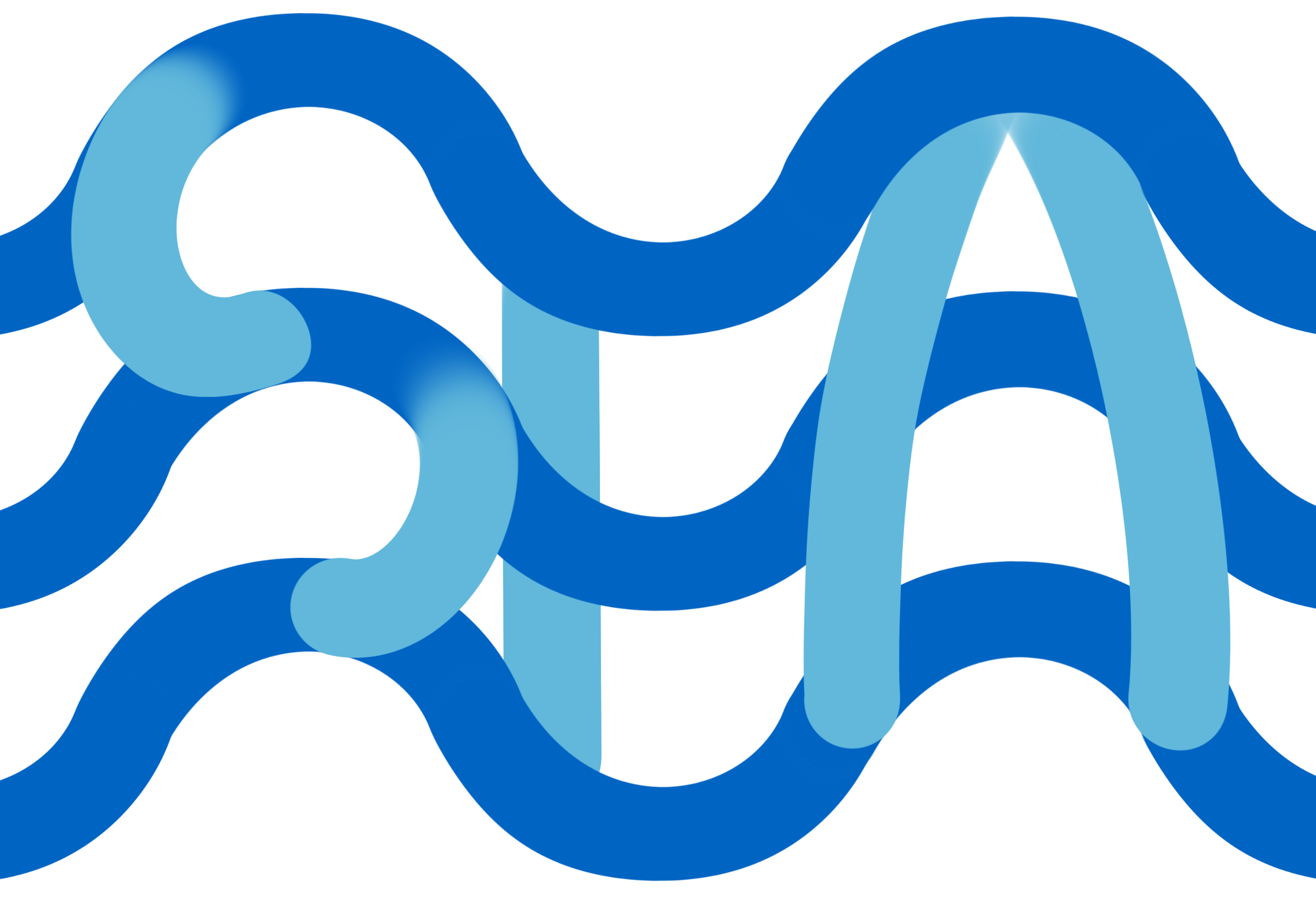} 
Automated Peer Reviewing in Paper SEA: {S}tandardization,
{E}valuation, 
and 
{A}nalysis
}

\author{Jianxiang Yu\textsuperscript{$\heartsuit$}\thanks{~~Equal Contribution},
Zichen Ding\textsuperscript{$\heartsuit$}\footnotemark[1], 
{\bf Jiaqi Tan\textsuperscript{$\heartsuit$}, }  
{\bf Kangyang Luo\textsuperscript{$\heartsuit$}, } 
{\bf Zhenmin Weng\textsuperscript{$\heartsuit$},} \\
{\bf Chenghua Gong\textsuperscript{$\heartsuit$},} 
{\bf Long Zeng\textsuperscript{$\heartsuit$},} 
{\bf Renjing Cui\textsuperscript{$\heartsuit$},} 
{\bf Chengcheng Han\textsuperscript{$\heartsuit$},}\\ 
{\bf Qiushi Sun\textsuperscript{$\diamondsuit$},}
{\bf Zhiyong Wu\textsuperscript{$\diamondsuit$},} 
{\bf Yunshi Lan\textsuperscript{$\heartsuit$},} 
{\bf Xiang Li\textsuperscript{$\heartsuit$}\thanks{~~Corresponding author}} \\
\textsuperscript{$\heartsuit$}
East China Normal University,
Shanghai, China \\
\textsuperscript{$\diamondsuit$} Shanghai AI Laboratory, 
Shanghai, China \\
\href{mailto:sea.ecnu@gmail.com}{sea.ecnu@gmail.com}
\\
\url{https://ecnu-sea.github.io/}
}

\begin{document}

\maketitle

\begin{abstract}

In recent years, the rapid increase in scientific papers has overwhelmed traditional review mechanisms, 
resulting in varying quality of publications. 
Although existing methods have explored the capabilities of Large Language Models~(LLMs) 
for
automated scientific reviewing,
their generated contents
are often generic or partial.
To address the issues above,
we introduce an automated paper reviewing framework SEA.
It comprises of three modules: Standardization, Evaluation, and Analysis, which are represented by models SEA-S, SEA-E, and SEA-A, respectively.
Initially, SEA-S distills data standardization capabilities of GPT-4 for integrating
multiple reviews for a paper.
Then, SEA-E utilizes standardized data for fine-tuning,
enabling it to generate constructive reviews.
Finally, SEA-A introduces a new evaluation metric called mismatch score to assess the consistency between paper contents and reviews. 
Moreover, we design a self-correction strategy to enhance the consistency.
Extensive experimental results on datasets collected from eight venues show that SEA can generate valuable insights for authors to improve their papers.

\end{abstract}

\input{sec-intro}
\input{sec-relatedworks}
\input{sec-method}
\input{sec-exp}

\input{sec-conclusion}

\input{sec-limitations}

\input{sec-ethics}

\section*{Acknowledgments}
This work is partially supported by the National Key R\&D Program of China (2023YFC3341200) and National Science Foundation of China (NSFC) (62137001).

\bibliography{custom}

\appendix

\input{sec-appendix}

\end{document}

%% file: sec-intro.tex
\section{Introduction}  
With the rapid pace of scientific advancement, there has been a significant increase in the volume of research publications
\cite{bornmann2015growth, reviewer2, lin2023automated}.
Nevertheless, it poses considerable challenges for traditional scientific feedback mechanisms~\cite{liang2023can}. 
On one hand, it exacerbates the pressure on the peer review process~\cite{lee2013bias, bjork2013publishing}; on the other hand, 
the disparate quality of these numerous publications can negatively affect the scientific research milieu~\cite{kelly2014peer, reviewergpt}. 
Consequently, there is a need for an
automated 
scientific reviewing
framework 
designed
to generate constructive reviews with strong evidence supports to  
help authors improve the caliber of their works~\cite{yuan2022can}.

\begin{figure}
    \centering
    \includegraphics[width=0.5\textwidth]{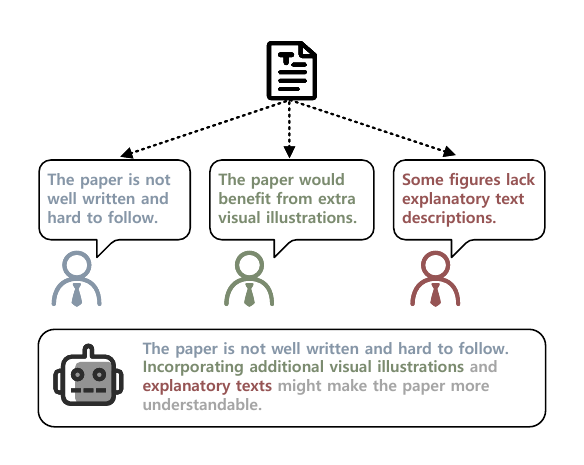}
    \caption{Multiple reviews of a paper often provide helpful but partial opinions on certain aspects. 
    Integrating these reviews can offer more comprehensive feedback on the paper.}
    \label{fig:intro1}
    \vspace{-1.2em}
\end{figure}

However, the task of delivering timely, thorough, and perceptive feedback on research papers is inherently intricate and {cognitively demanding}~\cite{horbach2018changing}.
Traditional language models typically struggle to handle such lengthy texts, let alone provide valuable review insights~\cite{cohan-etal-2020-specter, wang-etal-2020-reviewrobot}.
Fortunately, Large Language Models (LLMs) have demonstrated emergent capabilities~\cite{emergent},
which have shown
state-of-the-art performance in a wide range of tasks~\cite{brown2020language, touvron2023llama, tan2024llm,sun2024survey}.
Further,
they have also been strengthened to
handle increasingly longer contexts~\cite{jiang2023mistral},
facilitating the possibility for automated reviewing~\cite{liang2023can, reviewer2}.

Currently, some efforts have been made to explore the capabilities of LLMs for automated paper reviewing. 
For example, \citet{reviewergpt} and~\citet{liang2023can} investigate the potential reliability and credibility of paper reviews generated by LLMs with specially designed prompts.
Yet
most of these LLMs are tailored for broad and general-purpose applications~\cite{academicgpt}, 
so simply 
prompting 
LLMs in reviewing papers could output generic comments of less value~\cite{liang2023can}.
Further, 
certain studies have developed peer review datasets and fine-tuned LLMs to learn the paradigm of paper reviewing~\cite{academicgpt, reviewer2}.
However,
in the supervised fine-tuning (SFT) process,
these methods simply utilize a review for a paper that can be biased, partial~(see Figure~\ref{fig:intro1})
and often formalized in various formats and criteria,
which could hinder the potential of LLMs for automated paper reviewing~\cite{lin2023moprd, reviewer2}.
Also, they lack a self-correction mechanism when the generated reviews are less appealing.

\begin{figure*}
    \centering
    \includegraphics[width=1.0\linewidth]{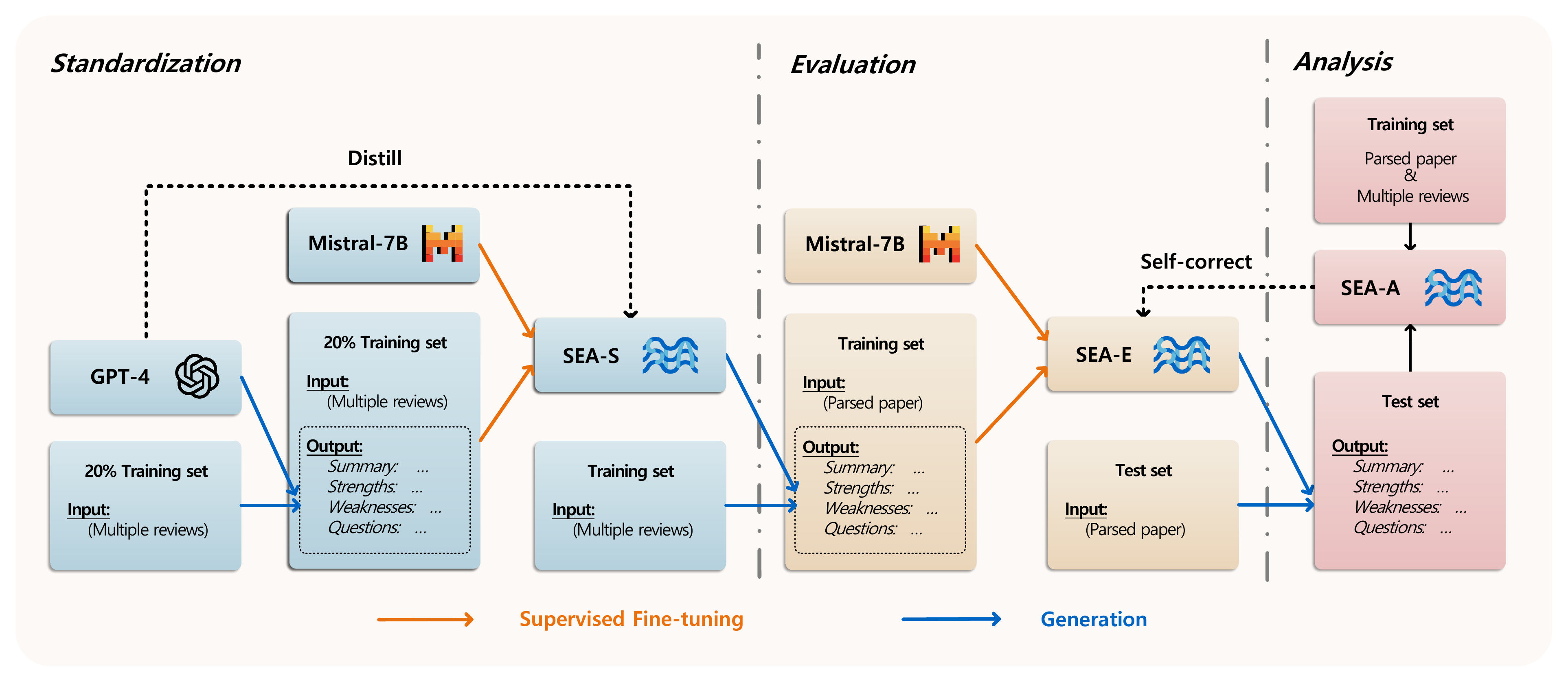}
    \caption{The overall framework of SEA consists of three modules:
    Standardization, Evaluation and Analysis.}
    \label{Framework:}
\end{figure*}

To tackle the issues, in this paper,
we 
propose a novel automated paper reviewing framework, namely, 
\textbf{SEA}, which consists of three modules: \textbf{\underline{S}tandardization},
\textbf{\underline{E}valuation}, and 
\textbf{\underline{A}nalysis}, as shown in Fig.~\ref{Framework:}.
We next summarize the details of each module.

In the \textbf{Standardization} module, 
we develop a 
model SEA-S, 
which aims to 
standardize reviews.
Specifically,
we first utilize 
GPT-4 to 
integrate multiple reviews of a paper into one that is in a unified format and 
criterion with constructive contents,
and form an instruction dataset for SFT.
After that,
we fine-tune an open-source LLM 
Mistral-7B
to distill the knowledge of GPT-4.

In the \textbf{Evaluation} module, we fine-tune another Mistral-7B to derive the SEA-E model,
which can comprehensively analyze papers and generate high-quality reviews.
Given papers that are in PDF format,
we parse them into text and LaTeX codes, and input their corresponding multiple reviews into SEA-S to generate standardized reviews.
The parsed papers, standardized reviews
and human-crafted prompts constitute another instruction dataset for SFT, leading to SEA-E.

In the \textbf{Analysis} module,
we further introduce a self-correction strategy
that promotes SEA to rethink and regenerate more constructive reviews,
when the generated reviews are inconsistent with the parsed papers.
To measure the inconsistency,
we put forward a metric, namely,
\emph{mismatch score}.
We also train a regression model SEA-A to estimate scores for the generated reviews.

Extensive experiments on eight diverse datasets show that the reviews generated by the SEA framework significantly outperform existing methods in terms of quality, comprehensiveness, and consistency.
To sum up, we highlight our contributions as follows:
\begin{itemize}[leftmargin=*, itemsep=0pt, parsep=0pt, topsep=0pt, partopsep=0pt]

\item We propose a novel framework SEA for automated paper reviewing.
\item We present an effective model SEA-S for standardizing reviews from various academic venues in different formats and criteria.

\item 
We devise 
a self-correction strategy to improve the consistency between papers and reviews.

\item 
We conduct 
extensive experiments to show the superiority of SEA over other competitors.

\end{itemize}

\emph{Finally, it is important to emphasize that the purpose of this paper is not to directly recommend the acceptance/rejection on papers. 
We anticipate our framework SEA can facilitate timely feedback for researchers, thereby enhancing the quality of their work and enabling them to transition efficiently to subsequent projects.}

%% file: sec-relatedworks.tex
\section{Related Works}
\subsection{Long-context Large Language Models}
LLMs have recently achieved substantial progress in accommodating lengthy contexts.
For example,
LongLLaMA~\cite{longllama} and LongLoRA~\cite{longlora} support long contexts processing by modifying the attention mechanism.
There are also  
some positional encoding methods proposed,
including ALiBi~\cite{alibi}, xPOS~\cite{xpos} and RoPE variants~\cite{chen2023extending, xiong2023effective}.

Assessing the capability of LLMs in handling long contexts has also attracted significant attention.
The needle-in-a-Haystack (NIAH) test~\cite{NIAH} has been widely adopted to evaluate long-context LLMs.  
Further, RULER~\cite{ruler} extends the vanilla NIAH test to
provide a more thorough assessment.
Based on the RULER evaluation results, we select Mistral-7B~\cite{jiang2023mistral} as the base model in our paper. 
Mistral-7B is a compact LLM that has been shown to handle at least 16K tokens, sufficient to meet the input requirements of most academic papers.

\subsection{Automated Scientific Reviewing}
Automating scientific reviewing 
began its investigation 
in the era of small language models.
The early work~\cite{zhang2022investigating} utilizes RoBERTa~\cite{roberta} to assess the textual fluency of papers and fairness disparity in peer review. 
In peer grading,
\citet{morris2023using} fine-tune distilBERT~\cite{distilbert} using course grading data from massive open online courses to examine the reliability of peer grading scores.
However, 
due to the restricted capability of language models in handling lengthy contexts,
automating scientific reviewing of a full paper has not been studied before the advent of LLMs.

Recently,
since LLMs exhibit advancements in various NLP tasks, 
some studies are exploring the capabilities of LLMs in automated paper reviewing. 
For example,
\citet{reviewergpt} and~\citet{liang2023can} customize prompts to guide GPT-4 in generating scientific feedbacks. 
\citet{academicgpt} conduct continuous training of LLaMA2-70B~\cite{touvron2023llama} on academic data, resulting in an academically enhanced model AcademicGPT. 
Further,
\citet{reviewer2} collect a large-scale peer review dataset, and propose a two-stage review generation framework REVIEWER2
with question-guided prompts.

%% file: sec-method.tex
\section{SEA}
This section details three major modules~(i.e., Standardization, Evaluation and Analysis) of SEA, and the overall framework is illustrated in Figure~\ref{Framework:}.

\subsection{SEA-S: Standardization}
To explore the potential of LLMs in automated scientific reviewing,
a high-quality labeled dataset is generally needed for supervised fine-tuning (SFT).
This process feeds LLMs with more peer reviews, 
thereby enhancing the quality of its generated ones. 
However, 
in the peer review datasets, 
each paper is often associated with multiple peer reviews, 
with each review offering a limited perspective based on the reviewer's field and expertise. 
On the other hand, the review formats and criteria could vary across different academic venues, and directly performing SFT on existing peer review datasets can lead to inconsistencies.
Therefore, we first have to standardize reviews in a unified format and criterion with comprehensive contents before SFT. 
For each paper, 
we integrate all the reviews into one,
which can eliminate redundancy and error in multiple reviews. 
The integrated review is expected to focus on the major advantages and disadvantages of the paper, thereby enhancing its quality.

To perform data standardization,
we attempt several representative open-source and closed-source models, such as Mistral-7B, GPT-3.5 and GPT-4. 
We empirically observe
that
Mistral-7B and GPT-3.5 tend to simply 
{concatenate the original contents}.
In contrast, 
GPT-4 leads them by
integrating reviews in an unified format and
providing detailed evidence for each argument (The comparative examples are given in Figure~\ref{fig:sea-s_exmaple} of Appendix~\ref{app:sea-s}). 
However, 
the API for GPT-4 is costly and inflexible.
Inspired by Alpaca~\cite{alpaca}, 
we distill GPT-4’s excellent data standardization capabilities into open-source models. 

Specifically, we first randomly select 20\% of the papers from the training set along with their reviews $\{[r^\text{origin}_{i1}, r^\text{origin}_{i2}, \dots, r^\text{origin}_{im}]\}^{n}_{i=1}$,
where $n$ is the number of {selected} papers and $m$ is the number of reviews corresponding to paper $p_i$.
Next, 
for each paper $p_i$,
we input all its reviews along with the customized instruction $inst_s$ into GPT-4, 
which in turn yields the standardized review $r^{\text{GPT-4}}_i$.
In this way, 
we can construct the \emph{instruction dataset} for the data standardization model SEA-S that takes Mistral-7B as the base model.
Formally,
the triplet
in the dataset is <$inst_s, [r^\text{origin}_{i1}, r^\text{origin}_{i2}, \dots, r^\text{origin}_{im}], r^{\text{GPT-4}}_i$>,
which is further served for SFT.
After fine-tuning SEA-S, 
we feed all the reviews in the training set into SEA-S for data standardization, 
which outputs the integrated reviews $\{r^\text{SEA-S}_{i}\}^{N}_{i=1}$. 
Here, $N$ denotes the number of papers in the training set.
In summary,
SEA-S provides a novel paradigm for integrating peer review data 
in an unified format across various conferences.

\subsection{SEA-E: Evaluation}
{In the {Evaluation} module, 
we aim to construct 
a talented LLM 
that can deeply understand papers
and generate constructive reviews.} 
Notably, 
since
raw crawled
papers are in PDF format, we first apply Nougat~\cite{nougat} as the parser, which is a model based on Visual Transformer and is specially designed for parsing academic documents.
In particular, 
Nougat can parse formulas into LaTeX codes instead of corrupted text encoding, enabling LLMs to gain a deeper understanding of papers' contents.
Further, 
due to the long-text characteristic of papers,
we choose the open-source model {Mistral-7B} as the backbone model, which has demonstrated its ability in 
effectively handling up to 16K tokens for the long-context benchmark RULER~\cite{ruler}.

Based on the outputs of the SEA-S model,
we next construct the \emph{instruction dataset} for the evaluation model SEA-E. 
Each triplet in the dataset is denoted as <$inst_e, \hat{p}_i,  r^\text{SEA-S}_i$>, 
where $inst_e$ is the specially designed instruction for evaluation, 
$\hat{p}_i$ is the parsed paper, 
and $r^\text{SEA-S}_i$ is the standardized review.
Note that
$r^\text{SEA-S}_i$ contains solid 
evidence for each argument in the review.
This endows SEA-E with
the capability to generate comprehensive and 
constructive 
reviews after SFT.

\subsection{SEA-A: Analysis}
Now, we step into the {Analysis} module,
where a \emph{mismatch score} is proposed to measure the consistency between papers and their generated reviews.
Given a paper $p$ with $m$ raw reviews,
let us denote its ground-truth \emph{paper ratings} as $S_p=\{s_{pr_1}, s_{pr_2}, \dots, s_{pr_m}\}$ and \emph{confidence scores} as $C_p=\{c_{pr_1}, c_{pr_2}, \dots, c_{pr_m}\}$, where
each $s_{pr_i}$ and $c_{pr_i}$ indicate the rating and confidence score given by the $i$-th reviewer.
We next use the confidence scores as weights and calculate the weighted average rating of paper $p$,
which is further subtracted from the reviewer’s rating to serve as the ground truth mismatch score. Formally, we have:
\begin{equation}
y_{true}^{pr_i} = s_{pr_i} - \frac{\sum_{j=1}^{m} c_{pr_j} * s_{pr_j} }{\sum_{j=1}^{m} c_{pr_j} }.
\end{equation}
From the equation, we see that,  when a reviewer’s rating is greater than the weighted average, the review may tend to emphasize the paper’s strengths; otherwise, the review may be preferably critical of the paper.
Generally,
the greater the difference, the lower the review quality.
When $y_{true}^{pr_i} = 0$, 
we consider the review to be relatively neutral and consistent with the paper content.
For example, when the review ratings of a paper are \{2, 6, 6, 6\} and all are given with full confidence, 
the quality of the review rated 2 is considered to be lower because it deviates significantly from the weighted average rating of 5. 

To estimate the mismatch score,
we train a lightweight regression model SEA-A.
Specifically,
each parsed paper $\hat{p}$ and its corresponding review $r$
generated from SEA-E
form a pair 
<{$\hat{p}$}, {$r$}>,
which serves as the input. 
We first utilize 
the pre-trained sentence representation model 
SFR-Embedding-Mistral~\cite{SFR} that is designed for long contexts to transform the texts of papers and reviews into representations $h_{\hat{p}}$ and $h_r$, respectively. 
Then, 
we 
compute the \emph{query} and \emph{key} vectors for 
both the paper and the review 
separately:
\begin{equation}
\begin{split}
    q_{\hat{p}} &= W^q h_{\hat{p}}, \quad   q_r = W^q h_r,\\
    k_{\hat{p}} &= W^k h_{\hat{p}}, \quad   k_r = W^k h_r.\\
\end{split}
\end{equation}
Here, 
$W^q$ and $W^k$ are learnable weight matrices. 
Based on
the query and key vectors,
we calculate the estimated mismatch score ${y}_{pred}^{pr}$
by:
\begin{equation}
    y_{pred}^{pr} = 
    w(q_{\hat{p}} {k_r}^T + q_r {k_{\hat{p}}}^T) + b.
\end{equation}
Finally, 
we use the mismatch score $y_{true}^{pr}$
as the ground truth
and the Mean Squared Error (MSE) loss as the objective to train the regression model SEA-A.
The smaller the absolute value of the mismatch score, 
the higher the consistency between the review and the paper.

\input{table/dataset}
\input{table/main_exp}

After SEA-A is trained, we further introduce 
a \emph{self-correction strategy} to analyze each review generated by SEA-E.
When the estimated mismatch score 
$y_{pred}^{pr}$ is larger than  
a pre-set threshold  $\theta$, 
we regenerate the review 
by adding the current mismatch score as additional prompt to ensure the consistency between the paper and the review.

%% file: table/dataset.tex

\begin{table*}[htbp]
  \centering
  \caption{Dataset Statistics}
  \resizebox{\linewidth}{!}{
    \begin{tabular}{rccccccccc}
    \toprule
          & \textbf{CONLL-16} & \textbf{ACL-17} & \textbf{COLING-20} & \textbf{ARR-22} & \textbf{NeurIPS-16-22} & \textbf{ICLR-17-23} & \textbf{NeurIPS-23} & \textbf{ICLR-24} & \textbf{Total} \\
    \midrule
    \# papers & 22    & 136   & 88    & 364   & 1,048 & 1,617 & 3,368 & 5,653 & 12,296 \\
    \# tokens per paper & 8,163 & 8,400 & 7,571 & 8,229 & 10,499 & 9,586 & 11,205 & 9,815 & 10,142 \\
    \# reviews & 39    & 272   & 112   & 684   & 3,847 & 5,779 & 15,027 & 21,839 & 47,602 \\
    \# tokens per review & 532   & 558   & 539   & 539   & 527   & 602   & 642   & 594   & 603  \\
    \% accepted & 50\%  & 67\%  & 93\%  & 100\% & 97\%  & 30\%  & 95\%  & 37\%  & 60\% \\
    domain & NLP/CL & NLP/CL & NLP/CL & NLP/CL & ML    & ML    & ML    & ML    & multi \\
    \bottomrule
    \end{tabular}%
    }
  \label{tab:dataset}%
\end{table*}

%% file: table/main_exp.tex
\begin{table*}[!htb]
\caption{The overall performance (\%) 
on four \textbf{cross-domain} datasets: 
CONLL-16, ACL-17, COLING-20, ARR-22, and four \textbf{in-domain} datasets: NeurIPS-16-22, ICLR-17-22, NeurIPS-23, ICLR-24.
We highlight the best score on each dataset in \textbf{bold} and the runner-up score with an \underline{underline}.
}
    \captionsetup{justification=centering}
    \begin{minipage}{.500\linewidth}
    
      \centering
        \resizebox{\linewidth}{!}{
    \begin{tabular}{l|c|ccc|ccc|c|c}
    \toprule
    \multicolumn{1}{c|}{\multirow{2}[2]{*}{Method}} & \multirow{2}[2]{*}{BLEU} & \multicolumn{3}{c|}{ROUGE (Recall)} & \multicolumn{3}{c|}{ROUGE (F1-score)} & \multirow{2}[2]{*}{BERTScore} & \multirow{2}[2]{*}{Tokens} \\
          &       & R-1   & R-2   & R-L   & R-1   & R-2   & R-L   &       &  \\
    \midrule
    \multicolumn{10}{c}{\textit{CONLL-16}} \\
    M-7B  & 18.92  & 20.81  & 4.81  & 10.30  & 28.66  & 6.81  & 14.18  & 82.49  & 554 \\
    M-7B-R & 18.16  & 21.96  & 5.17  & 10.62  & 29.56  & 7.18  & 14.31  & 82.57  & 357 \\
    M-7B-3.5 & 19.70  & 26.51  & 5.58  & 13.96  & 30.19  & 6.45  & 15.37  & 82.01  & 627 \\
    SEA-E & \underline{29.07} & \underline{34.91} & \underline{7.79} & \underline{15.29} & \underline{38.64} & \underline{8.67} & \underline{16.73} & \underline{82.85} & 793 \\
    SEA-EA & \textbf{31.01} & \textbf{36.96} & \textbf{8.91} & \textbf{16.34} & \textbf{40.49} & \textbf{9.68} & \textbf{17.57} & \textbf{82.94} & 798 \\
    \midrule
    \multicolumn{10}{c}{\textit{ACL-17}} \\
    M-7B  & 18.92  & 21.53  & 5.23  & 10.50  & 27.99  & 6.93  & 13.54  & 82.75  & 569 \\
    M-7B-R & 18.15  & 21.84  & 5.19  & 10.76  & 27.71  & 6.87  & 13.55  & 82.56  & 357 \\
    M-7B-3.5 & 16.73  & 27.27  & 6.26  & 14.47  & 26.09  & 6.19  & 13.19  & 82.37  & 636 \\
    SEA-E & \underline{25.67} & \underline{33.13} & \underline{7.71}  & \underline{14.94} & \underline{35.52} & \underline{8.45}  & \underline{15.62} & \underline{83.08}  & 772 \\
    SEA-EA & \textbf{27.90} & \textbf{35.83} & \textbf{8.84} & \textbf{15.83} & \textbf{38.03} & \textbf{9.48} & \textbf{16.36} & \textbf{83.19} & 806 \\
    \midrule
    \multicolumn{10}{c}{\textit{COLING-20}} \\
    M-7B  & 21.97  & 29.11  & 6.42  & 14.80  & 31.91  & 7.01  & 15.83  & 82.76  & 579 \\
    M-7B-R & 19.49  & 29.21  & 6.69  & 15.20  & 30.23  & 6.80  & 15.25  & 82.27  & 361 \\
    M-7B-3.5 & 18.13  & 34.03  & 7.56  & 18.43  & 28.49  & 6.10  & 14.77  & 82.12  & 617 \\
    SEA-E & \underline{22.93} & \underline{40.62} & \underline{9.23} & \underline{20.05} & \underline{34.37} & \underline{7.65} & \underline{16.15} & \underline{82.85} & 774 \\
    SEA-EA & \textbf{24.85} & \textbf{42.97} & \textbf{10.57} & \textbf{20.89} & \textbf{36.67} & \textbf{8.76} & \textbf{16.96} & \textbf{83.09} & 782 \\
    \midrule
    \multicolumn{10}{c}{\textit{ARR-22}} \\
    M-7B  & 22.07  & 25.28  & 6.96  & 12.46  & 32.60  & 9.16  & 15.99  & 83.25  & 575 \\
    M-7B-R & 20.27  & 24.89  & 6.70  & 12.60  & 31.22  & 8.66  & 15.71  & 82.70  & 357 \\
    M-7B-3.5 & 20.18  & 31.70  & 7.90  & 16.38  & 30.82  & 7.86  & 15.33  & 82.65  & 650 \\
    SEA-E & \underline{27.92} & \underline{37.64} & \underline{9.37} & \underline{17.18} & \underline{38.94} & \underline{9.84} & \underline{17.35} & \underline{83.38} & 787 \\
    SEA-EA & \textbf{30.05} & \textbf{40.34} & \textbf{10.82} & \textbf{18.17} & \textbf{41.37} & \textbf{11.19} & \textbf{18.20} & \textbf{83.59} & 818 \\
    \bottomrule
    \end{tabular}%
}
    
    \end{minipage}%
    \hfill
    \begin{minipage}{.495\linewidth}
      \centering
        \resizebox{\linewidth}{!}{
    \begin{tabular}{l|c|ccc|ccc|c|c}
    \toprule
    \multicolumn{1}{c|}{\multirow{2}[2]{*}{Method}} & \multirow{2}[2]{*}{BLEU} & \multicolumn{3}{c|}{ROUGE (Recall)} & \multicolumn{3}{c|}{ROUGE (F1-score)} & \multirow{2}[2]{*}{BERTScore} & \multirow{2}[2]{*}{Tokens} \\
          &       & R-1   & R-2   & R-L   & R-1   & R-2   & R-L   &       &  \\
    \midrule
    \multicolumn{9}{c}{\textit{NeurIPS-16-22}}                            &  \\
    M-7B  & 14.91  & 14.47  & 4.89  & 7.15  & 23.31  & 7.94  & 11.56  & 83.10  & 612 \\
    M-7B-R & 13.94  & 14.47  & 4.79  & 7.29  & 22.70  & 7.67  & 11.44  & 82.73  & 362 \\
    M-7B-3.5 & 16.95  & 20.41  & 6.02  & 10.72  & 26.45  & 8.13  & 13.45  & 82.56  & 629 \\
    SEA-E & \underline{24.83} & \underline{24.12} & \underline{7.31} & \underline{10.66} & \underline{34.06} & \underline{10.44} & \underline{15.11} & \underline{83.35} & 782 \\
    SEA-EA & \textbf{27.08} & \textbf{26.76} & \textbf{8.38} & \textbf{11.55} & \textbf{36.91} & \textbf{11.69} & \textbf{15.99} & \textbf{83.52} & 838 \\
    \midrule
    \multicolumn{9}{c}{\textit{ICLR-17-23 }}                              &  \\
    M-7B  & 13.75  & 13.10  & 4.42  & 6.51  & 21.65  & 7.36  & 10.80  & 83.26  & 607 \\
    M-7B-R & 12.98  & 13.38  & 4.45  & 6.85  & 21.36  & 7.26  & 10.91  & 82.80  & 359 \\
    M-7B-3.5 & 17.85  & 18.26  & 5.70  & 9.27  & 27.37  & 8.69  & 13.94  & 82.87  & 637 \\
    SEA-E & \underline{23.34} & \underline{22.38} & \underline{6.84} & \underline{9.93} & \underline{32.50} & \underline{10.07} & \underline{14.49} & \underline{83.58}  & 783 \\
    SEA-EA & \textbf{25.47} & \textbf{24.80} & \textbf{7.87} & \textbf{10.81} & \textbf{35.23} & \textbf{11.32} & \textbf{15.43} & \textbf{83.73} & 841 \\
    \midrule
    \multicolumn{9}{c}{\textit{NeurIPS-23}}                               &  \\
    M-7B  & 12.42  & 11.96  & 4.96  & 6.13  & 20.55  & 8.55  & 10.55  & 83.86  & 617 \\
    M-7B-R & 11.92  & 11.88  & 4.87  & 6.16  & 20.14  & 8.31  & 10.49  & 83.44  & 366 \\
    M-7B-3.5 & 16.71  & 16.80  & 6.12  & 8.53  & 26.51  & 9.74  & 13.50  & 83.20  & 650 \\
    SEA-E & \underline{21.34} & \underline{20.32} & \underline{7.27} & \underline{9.14} & \underline{31.34} & \underline{11.26} & \underline{14.14} & \underline{84.02}  & 794 \\
    SEA-EA & \textbf{23.32} & \textbf{22.49} & \textbf{8.38} & \textbf{9.91} & \textbf{34.03} & \textbf{12.73} & \textbf{15.03} & \textbf{84.20} & 844 \\
    \midrule
    \multicolumn{9}{c}{\textit{ICLR-24}}                                  &  \\
    M-7B  & 13.93  & 13.48  & 5.29  & 6.73  & 22.55  & 8.89  & 11.28  & 83.79  & 614 \\
    M-7B-R & 13.91  & 14.17  & 5.41  & 7.21  & 22.94  & 8.85  & 11.69  & 83.81  & 380 \\
    M-7B-3.5 & 18.72  & 19.40  & 6.52  & 9.64  & 29.26  & 9.93  & 14.58  & 83.29  & 649 \\
    SEA-E & \underline{23.88} & \underline{23.28} & \underline{7.90} & \underline{10.13} & \underline{34.29} & \underline{11.71} & \underline{14.98} & \underline{84.04} & 793 \\
    SEA-EA & \textbf{25.96} & \textbf{25.62} & \textbf{8.97} & \textbf{10.97} & \textbf{36.97} & \textbf{13.02} & \textbf{15.88} & \textbf{84.15} & 852 \\
    \bottomrule
    \end{tabular}%

        }
    \end{minipage}
        
    \label{tab:main_result}
\end{table*}

%% file: sec-exp.tex
\section{Experiments}
\label{Experiments:sec}

\subsection{Experimental Details}
\paragraph{Datasets.}
We crawl the latest papers and their corresponding reviews from OpenReview
\footnote{\url{https://openreview.net/}}, 
including NeurIPS-2023 and ICLR-2024. 
We randomly sample 90\% of the data according to the distribution of ``Rating'' to serve as the training set, with the remaining 10\% used as the test set for evaluation.
Our test set {also} includes subsets from REVIEWER2~\cite{reviewer2} for NeurIPS (2016-2022) and ICLR (2017-2023).
Additionally, we conduct evaluations on cross-domain datasets from Natural Language Processing~(NLP) and Computational Linguistics~(CL) fields,
incorporating data from PeerRead~\cite{peerread} for CONLL-2016 and ACL-2017,
and from NLPeer~\cite{dycke2022nlpeer} for COLING-2020 and ARR-2022.
All the datasets include the original PDF files of the papers and structurally formatted reviews.
Different review data exhibits format difference across various conferences and years.
The statistics of our datasets are summarized
in Table~\ref{tab:dataset}.

\paragraph{Setup.}
We 
use \emph{Mistral-7B-Instruct-v0.2}~\cite{jiang2023mistral}
with a context length of 32k
as our backbone model.
In the Evaluation module,
the reviews that our methods generate consists of three parts: a textual part with ``Summary'', ``Strengths'', ``Weaknesses'', and ``Questions''; 
a quantitative part that includes ``Soundness'', ``Presentation'', ``Contribution'', and ``Rating''; 
and finally,
the paper decision~(Accept/Reject) 
with corresponding reasons.
In the Analysis module,
we utilize 80\% of the entire training set for training and the remaining 20\% for validation.
We set the threshold $\theta$ to the average mismatch score in the validation set.
In our framework, there are two methods for generating reviews: \textbf{SEA-E} and \textbf{SEA-EA}, where
SEA-EA is an enhanced model that combines the Analysis module with SEA-E.
For SEA-EA,
if the mismatch score between  generated reviews and papers surpasses $\theta$, 
this score will be incorporated into the prompts to improve the quality of generated reviews.
Moreover,
if the mismatch score consistently exceeds $\theta$ across 10 successive trials,
the generation process will be terminated.
The review with the smallest score will be selected as the final output.

\paragraph{Baselines.}
We compare the following baseline methods,
which are divided into two categories:
(1) Direct inference with LLMs: 
We directly use \textbf{M}istral-\textbf{7B} (\textbf{M-7B}) for inference, guided by $inst_e$ to generate reviews in the specified format.
(2) SFT methods: 
From all reviews for each paper in the training set, 
we randomly select one review as the output for SFT, 
referred to as \textbf{M}istral-\textbf{7B}-\textbf{R}andom (\textbf{M-7B-R}). 
\textbf{M}istral-\textbf{7B}-GPT-\textbf{3.5} (\textbf{M-7B-3.5}) refers to the method where reviews for each paper are standardized using $\emph{gpt-3.5-turbo}$, and these standardized outputs are then applied in the SFT stage.
Moreover, 
REVIEWER2~\cite{reviewer2} is a two-stage
review generation framework.
Due to time-consuming,
we use a smaller test set and compare it with REVIEWER2. 
The detailed experimental results are provided in the Appendix~\ref{app:reviewer2}.

We unify the instruction $inst_e$ and input $\hat{p}$ across all the baseline methods and our framework. 
Here, $inst_e$ is the instruction for SEA-E, and $\hat{p}$ represents the parsed paper.
Detailed information about $inst_e$ can be found in Table~\ref{tab:instruction_sea-s} in Appendix~\ref{app:sea-s}.

\subsection{Main Results}
\label{sec:4.2}


We use BLEU~\cite{papineni2002bleu}, ROUGE (Recall), ROUGE (F1-score)~\cite{lin2004rouge}, 
and BERTScore~\cite{bertscore} 
as metrics to evaluate the quality of generated reviews across eight datasets. 
Specifically,
BLEU and ROUGE measure the similarity between papers and reviews based on n-grams,
while BERTScore focuses on semantic similarity in the embedding space.
For the ROUGE metric, 
recall measures how comprehensively the generated reviews capture the key information from raw papers, 
while the F1 score assesses the balance between precision and recall in the generated contents.
To measure the completeness and comprehensiveness of the generated reviews,
we simply concatenate all {the} reviews of each paper to serve as a benchmark for evaluation.
Moreover, 
we have also counted the average number of tokens in the generated reviews.

The results in Table~\ref{tab:main_result} show that SEA outperforms other baseline models across all the testing scenarios, 
with particularly notable gains on the ROUGE (Recall) metric.
This confirms that our proposed framework SEA is capable of generating comprehensive and constructive reviews.
Further, 
SEA not only performs excellently on in-domain tasks but also shows strong performance on cross-domain datasets, 
demonstrating its robust generalizability.
It is also worth noting that SEA-EA surpasses SEA-E in all cases,
underscoring the effectiveness of the self-correction strategy in
generating well-grounded reviews consistent with raw papers.
However, 
for M-7B-R, we notice that randomly selecting a review as the output of SFT often leads to shorter texts.
To some extent,
the quality of a review is positively correlated with its length, which explains its poor performance.
Although directly inferring with M-7B can
generate longer text, 
it fails to align with human reviews, resulting in lower evaluation scores.
For M-7B-3.5, its performance is poorer than SEA-E,
which {further} indicates the effectiveness of SEA-S. 
Consequently,
using high-quality standardized data generated by SEA-S can effectively improve the performance of SFT.
In Appendix~\ref{app:sea-e} we give concrete examples of reviews
generated by different models.

\subsection{Comparison of Standardized Results}
\begin{figure}[h]
    \centering
    \includegraphics[width=1.0\linewidth]{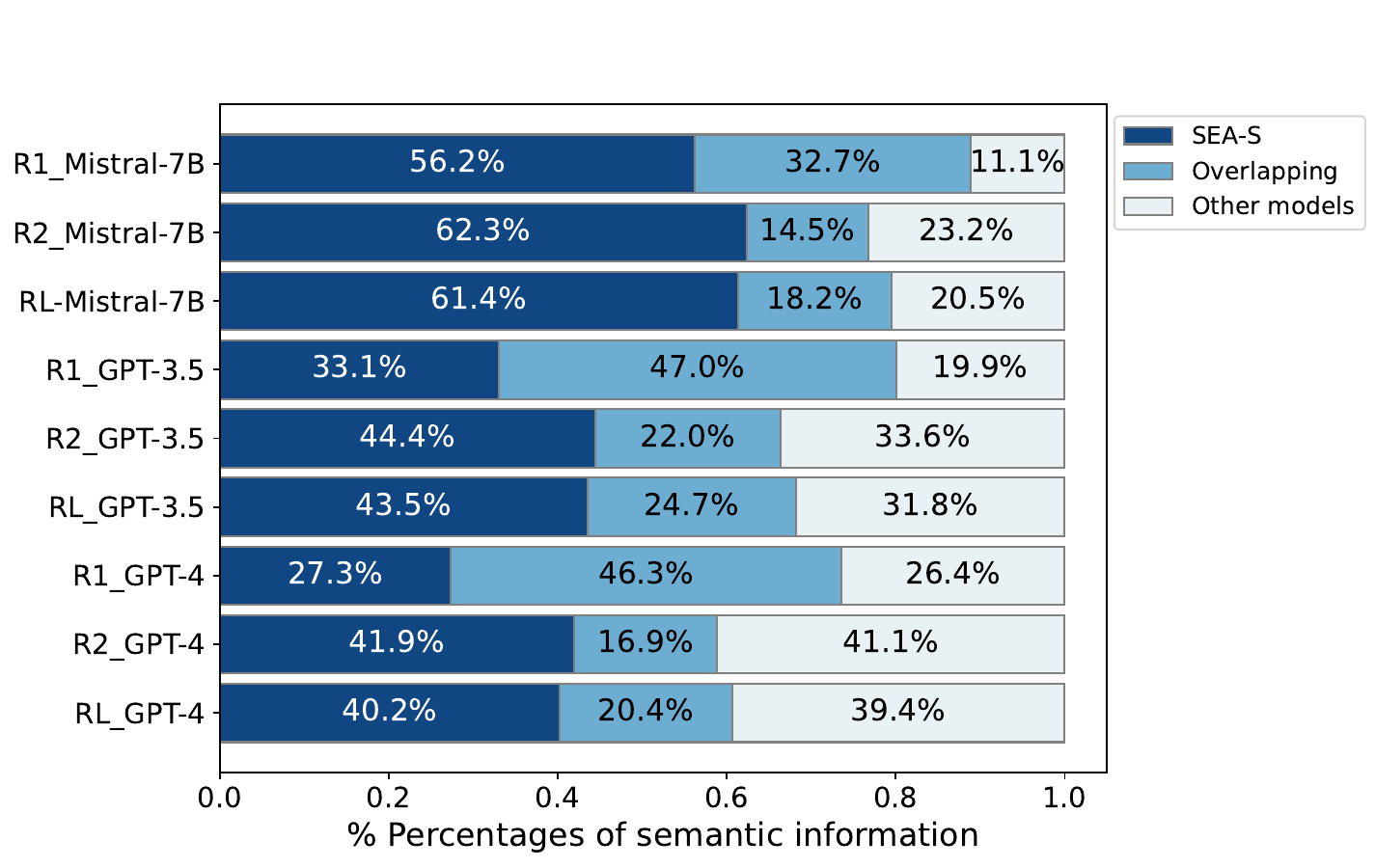}
    \caption{
    Content analysis results.
    }
    \label{fig:sea-s}
\end{figure}

We 
show the standardized results on  
papers in the training set of NeurIPS-2023 and ICLR-2024 that have different rating criteria. In addition, reviews are organized in various formats.
\paragraph{Content analysis.}
We first compare SEA-S with Mistral-7B, GPT-3.5, and GPT-4 to evaluate their review standardization performance. 
All the models are fed with the same inputs, including the instruction $inst_s$ and multiple reviews.
Since there is no ground-truth text for this standardized task,  
we utilize reviews generated by SEA-S as \emph{references}, 
while reviews generated by other models serve as \emph{candidates}.
Next, we calculate \emph{recall} and \emph{precision} values of ROUGE for candidates compared to references.
Based on
the content intersection of reference and candidate,
{recall} and precision refer to the percentage of intersection
in reference and candidate, respectively. 
From the two metrics,
we can deduce
the percentages of overlapping and exclusive semantic information 
in both reviews,
whose results are shown in Figure~\ref{fig:sea-s}.
We compare the model performance 
w.r.t. different ROUGE metrics, including ROUGE-1~(R1), ROUGE-2~(R2), and ROUGE-L~(RL). 
The light blue area in the figure indicates the overlapping contents, while the dark blue and light grey areas represent the exclusive contents by SEA-S (reference) and other models (candidate), respectively.

From the figure, we see that,
SEA-S can generate a significantly larger percentage of exclusive contents than both Mistral-7B and GPT-3.5.
This further verifies that SEA-S can
better standardize reviews with richer information. 
We also surprisingly observe that 
SEA-S can output slightly more exclusive contents in standardized reviews
than GPT-4.
The reason could be that 
the instruction dataset for SFT in SEA-S is derived from GPT-4.
Considering the high cost of GPT-4, 
this demonstrates the effectiveness of small models for review standardization.
On the other hand, 
recap that the difference between M-7B-3.5 and SEA-E only lies in the data standardization step.
The advantage of SEA-E over 
M-7B-3.5 in 
Table~\ref{tab:main_result} shows 
that SEA-S has better data standardization capability.

\begin{figure}[h]
    \centering
\includegraphics[width=0.85\linewidth]{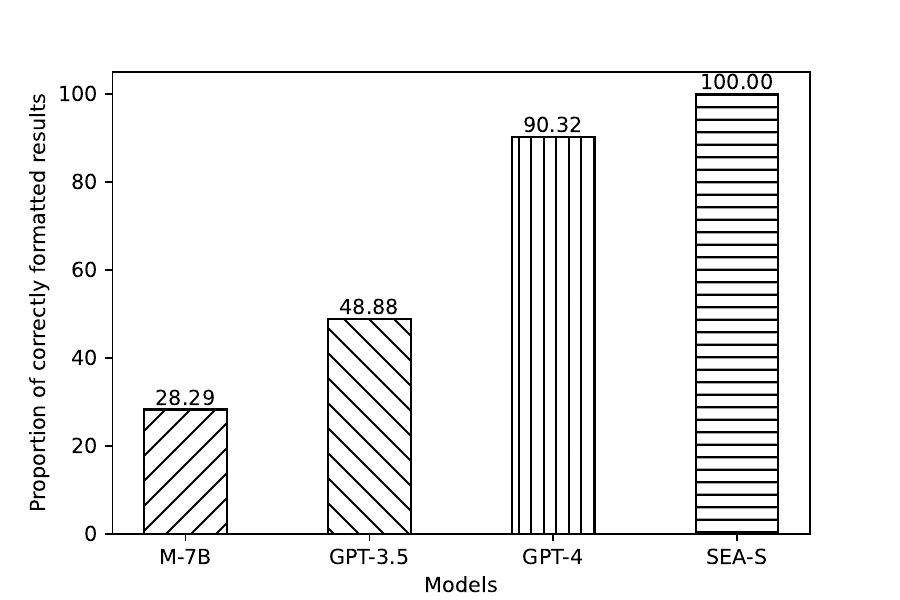}
    \caption{Format analysis of different models.}
    \label{fig:sea-s_format}
\end{figure}

\paragraph{Format analysis.}
{Standardized data} formats can help LLMs better understand the correspondence between the instruction and generated content during SFT.
To perform format analysis, 
we utilize regular expression matching based on instruction formats to calculate the proportion of correctly formatted reviews integrated by different models.
The results given in
Figure~\ref{fig:sea-s_format} demonstrate that SEA-S is capable of generating 100\% correctly formatted data.
In contrast, Mistral-7B and GPT-3.5 show poor performance, particularly the former, which generates a large amount of data that does not meet the format requirements.
Also, we observe that around 10\% of the data integrated by GPT-4 does not fully comply with the instruction.
Compared to GPT-4, SEA-S benefits from SFT and thus shows superior instruction adherence.
Overall, SEA-S demonstrates excellent effectiveness in handling reviews of various formats and criteria.
Details of the instruction for standardizing reviews and specific examples are given in Appendix~\ref{app:sea-s}.

\subsection{Mismatch Score in \modelname-A}
\begin{figure}[!h]
    \centering
\includegraphics[width=1.0\linewidth]{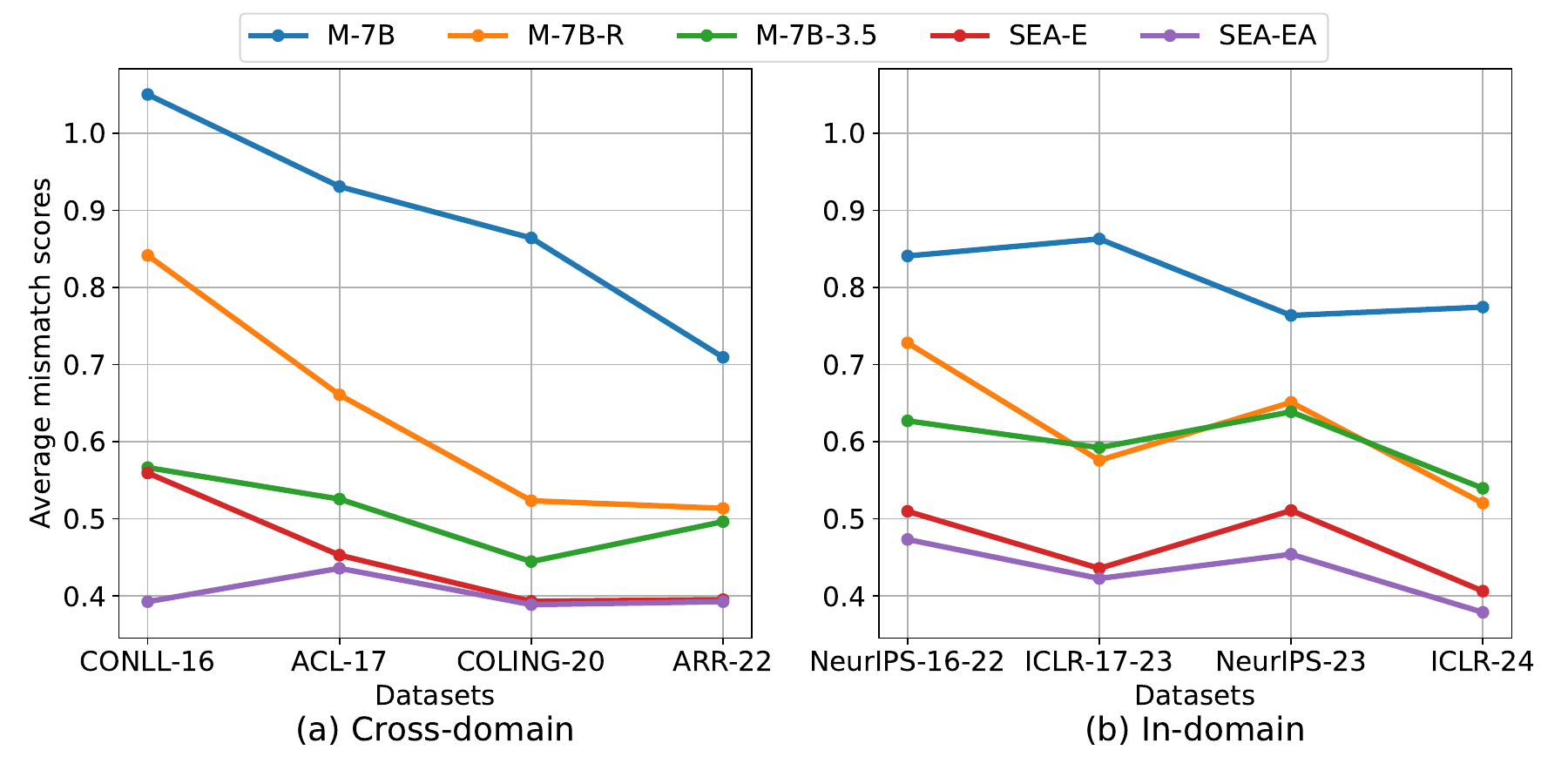}
    \caption{The performance of different models on mismatch scores across various datasets.}
    \label{fig:mms}
\end{figure}

To analyse the consistency between the reviews generated by different models and the corresponding papers, 
we input the reviews and their respective papers in the test set into the trained SEA-A model to 
calculate the average mismatch score for each model across different datasets. 
As illustrated in Figure~\ref{fig:mms},
SEA-EA, due to its self-correction strategy,
consistently outperforms others
across all the datasets.
Further,
SEA-E is the runner-up method.
This verifies that the reviews generated by both methods have a higher consistency with their corresponding papers. Mistral-7B, which has not undergone fine-tuning, 
fails to learn the correspondence between papers and reviews, resulting in higher mismatch scores. 
Although M-7B-R and M-7B-3.5 are fine-tuned, 
they are still worse than our methods.
This can be explained by 
the insufficient model standardization capability.

To further study mismatch score,
for each paper,
we randomly select a review from other papers in the test set as the ``negative'' review.
The negative review is expected to derive a larger mismatch score than the generated review, which is empirically observed from our results given in 
Appendix~\ref{app:sea-a}.
This again confirms that our regression model is capable of quantitatively assessing the consistency across reviews and papers.

\subsection{Quantitative Score Analysis}
We conduct a further quantitative analysis on the four scores 
in the generated reviews on two datasets with actual scores, NeurlPS-2023 and ICLR-2024.
The four scores include ``Soundness'', ``Presentation'', and ``Contribution'', 
which are integers from $[1,4]$, and ``Rating'', 
which is an integer from $[1,10]$.
The rating criterion is given in the instruction of SEA-E in Table~\ref{tab:instruction_sea-e}. 
In practice,
each paper has multiple reviews and each review has the above four scores.
Therefore, 
given a paper,
for each score,
we use the ``Confidence'' score in each review as the weight and calculate the weighted average as the reference score.

\input{table/score}

To assess the discrepancy between the generated scores and the reference scores, 
we use the Mean Squared Error (MSE) metric.
The lower the MSE value, 
the more accurate the generated results.
In Table~\ref{tab:score}, 
the percentages in parentheses indicate the proportions of generated reviews with valid scores,
while ``N/A'' denotes those with unsuccessful generations~(e.g. text is generated instead of scores).
It can be seen that our proposed method ensures the validity of the output format, whereas other models tend to generate content that does not comply with the instruction to varying degrees,
especially M-7B that has not undergone SFT.
The MSE metric shows
that our proposed methods outperform the baseline models in practically all cases.
Although SEA-E scores larger than M-7B-3.5 by  0.02 in the ``Presentation'' on ICLR-2024, 
SEA-E achieves 100\% valid scores in generation, whereas M-7B-3.5 only reaches 86\%.
Additionally, 
SEA-EA demonstrates improvements over SEA-E in most cases, 
further validating that a self-correcting strategy allows for high consistency between generated results and human feedback on quantitative evaluation results.

\subsection{Qualitative Decision Analysis}

In this part,
we analyze ``Decision'' and ``Reason'' of the generated review,
i.e., the final decision (accept or reject) of the paper and the corresponding reasons.
Typically, the Area Chair (AC) gives the final decision and meta-reviews.
We calculate the accuracy, precision, recall, and F1-score of the generated results compared to the final decisions, 
and use BERTScore to measure the semantic similarity between the reasons and meta-reviews.
The model
M-7B-R randomly selects a review for SFT that does not include the decision or the meta reviews, hence 
we do not take it as baseline.

\input{table/decision}

From Table~\ref{tab:decision}, 
it can be seen that 
SEA-EA leads to the largest accuracy and 
BERTScore values,
where the latter
shows the model's effectiveness in generating
reasons semantically aligned with meta-reviews.
Due to the acceptance rate of 95\% in the NeurIPS-2023 test set~(see Table~\ref{tab:dataset}), the overall results are large. 
For ICLR-2024, 
the accuracy of SEA-EA surpasses that of SEA-E over 4\%, further indicating the effectiveness of 
the self-correction strategy.
Additionally, we note that M-7B exhibits high recall about 97\%, but poor precision, suggesting a tendency to cater to human preferences by accepting most papers.
In contrast, 
our method performs better in both Precision and F1-score, which indicates that ours
can identify papers of different quality more effectively.
Overall, SEA aligns more closely with actual AC decisions and refrains from favoring decisions that lean towards acceptance.

\subsection{Human and GPT evaluation}

To further validate the performance of SEA, we supplemented the study with a questionnaire experiment for SEA-E and SEA-S, evaluated by both humans and GPT. 
The specific content of the questionnaire can be found in Appendix~\ref{app:questionnair}.

\subsubsection{Review Quality~(SEA-E)}

Referring to previous review quality instruments~\cite{van1999development}, we design 10 questions across different aspects, with each question rated on a scale of 1 to 10, where a higher score indicates better review quality. 
We then randomly sample 20 papers and evaluate the review quality generated by different models.

\paragraph{Human evaluation.}

We invite 20 qualified experts to evaluate reviews generated by various models, with each expert randomly assigned to assess 5 papers. 
Additionally, we include an extra question in the evaluation process: after anonymizing the model names, experts are asked to select the best review for each paper. 
The evaluation results from each expert were then aggregated, and the mean scores were calculated.

\paragraph{GPT evaluation.}

Given the inherent subjectivity in human evaluations and inspired by the work of ~\citet{zheng2023judging}, we employ $\emph{gpt-4o-2024-05-13}$ to score the reviews generated by different models using the same set of 10 questions. 
Due to the limitation of context length, GPT-4o did not perform the top-1 ranking evaluation.

\input{table/human_eval_sea_e}

From the Table~\ref{tab:sea_e_eval}, it can be seen that SEA-E performs exceptionally well in both GPT-4o and human evaluations, significantly outperforming other models on ten questions. 
Additionally, in 90\% of the cases, SEA-E was preferred by the experts.

\subsubsection{Standardized Content Quality (SEA-S)}


We design seven questions to assess the standardized content across different models, using the same scoring criteria for evaluating review quality. 
Then, we randomly select 10 papers for evaluation.

\paragraph{Human evaluation.}

We invite 10 experts to evaluate the standardized content generated by different models, with each expert being randomly assigned 5 papers for assessment.

\paragraph{GPT evaluation.}

We adopt the $\emph{GPT-4o}$ to score the standardised content generated by the different models based on the same questionnaire.

\input{table/human_eval_sea_s}



Table~\ref{tab:sea_s_eval} presents the results from two distinct evaluation methods. 
The human evaluation emphasizes SEA-S's strengths, especially in Q4, Q5, and Q6, where it excels in content relevance, logical coherence, and conciseness. 
While GPT-4o generally assigns higher scores across the board, SEA-S consistently stands out compared to other models.

Based on the results of the questionnaire, 
the effectiveness of the SEA framework is further confirmed. 
It demonstrates the ability to generate standardized content, leading to comprehensive and high-quality review feedback.

%% file: table/score.tex
\begin{table}[!htb]
  \centering
  \caption{Quantitative Score Analysis.}
  \resizebox{\linewidth}{!}{
    \begin{tabular}{clcccc}
    \toprule
          & Method & Soundness & Presentation & Contribution & Rating \\
    \midrule
    \multirow{5}[2]{*}{\begin{sideways}NeurIPS-23\end{sideways}} & M-7B  & N/A   & N/A   & N/A   & 8.51~(10\%) \\
          & M-7B-R & 0.20~(99\%) & 0.26~(99\%) & 0.32~(99\%) & 1.44~(99\%) \\
          & M-7B-3.5 & 0.15~(99\%) & 0.16~(99\%) & 0.27~(99\%) & 1.14~(99\%) \\
          & SEA-E & \underline{0.12~(100\%)} & \textbf{0.14~(100\%)} & \underline{0.18~(100\%)} & \underline{0.80~(100\%)} \\
          & SEA-EA & \textbf{0.11~(100\%)} & \underline{0.15~(100\%)} & \textbf{0.17~(100\%)} & \textbf{0.73~(100\%)} \\
    \midrule
    \multirow{5}[2]{*}{\begin{sideways}ICLR-24\end{sideways}} & M-7B  & N/A   & N/A   & N/A   & 12.96~(13\%) \\
          & M-7B-R & 0.32~(99\%) & 0.39~(99\%) & 0.42~(99\%) & 2.12~(99\%) \\
          & M-7B-3.5 & 0.32~(86\%) & \underline{0.28~(86\%)} & 0.45~(86\%) & 2.50~(86\%) \\
          & SEA-E & \underline{0.28~(100\%)} & 0.30~(100\%) & \underline{0.38~(100\%)} & \underline{2.11~(100\%)} \\
          & SEA-EA & \textbf{0.27~(100\%)} & \textbf{0.24~(100\%)} & \textbf{0.34~(100\%)} & \textbf{1.72~(100\%)} \\
    \bottomrule
    \end{tabular}%
}
  \label{tab:score}%
\end{table}%

%% file: table/decision.tex


\begin{table}[!htb]
  \centering
  \caption{Qualitative Decision Analysis. 
  The symbol (*) indicates that there are incompleteness or errors in the generated content;
  only valid generations are counted.}
  \resizebox{\linewidth}{!}{
  \begin{tabular}{clcccccc}
    \toprule
          & Method & Accuracy & Precision & Recall & F1-score & BERTScore \\
    \midrule
    \multirow{4}[2]{*}{\begin{sideways}NeurIPS-23\end{sideways}} & M-7B* & 93.18 & 94.01  & 99.05  & 96.47  & 84.27  \\
          & M-7B-3.5* & 81.01 & 95.34  & 83.91  & 89.26  & 84.04  \\
          & SEA-E & 99.41 & 99.37  & \textbf{100.0} & 99.69  & 84.21  \\
          & SEA-EA & \textbf{99.70} & \textbf{99.69}  & \textbf{100.0} & \textbf{99.84}  & \textbf{85.22}  \\
    \midrule
    \multirow{4}[2]{*}{\begin{sideways}ICLR-24\end{sideways}} & M-7B* & 36.81 & 37.14  & \textbf{97.65}  & 53.82  & 84.19  \\
          & M-7B-3.5* & 50.27 & 39.63  & 61.03  & 48.06  & 84.61  \\
          & SEA-E & 54.16 & 43.31  & 69.95  & 53.50  & 85.07  \\
          & SEA-EA & \textbf{58.23} & \textbf{46.48}  & 71.36  & \textbf{56.30}  & \textbf{86.08}  \\
    \bottomrule
    \end{tabular}%
  }
  \label{tab:decision}%
\end{table}%

%% file: table/human_eval_sea_e.tex
\begin{table}[htbp]
\Large
\centering
\caption{Review quality evaluation by humans and GPT.~(`R2' refers to REVIEWER2.)}
\resizebox{\linewidth}{!}{
\begin{tabular}{c|c|c|c|c|c|c|c|c}
\toprule
\multirow{2}{*}{} & \multicolumn{4}{c|}{Human Evaluation} & \multicolumn{4}{c}{GPT Evaluation} \\ \cline{2-9} 
                        & \normalsize M-7B-R & \normalsize M-7B-3.5 & \normalsize R2   & \normalsize SEA-E        & \normalsize M-7B-R  & \normalsize M-7B-3.5 & \normalsize R2    & \normalsize SEA-E    \\ \hline
Q1                       & 5.1   & 6.2     & 3.8 & \textbf{7.9} & 5.0    & 6.5    & 4.5  & \textbf{7.2} \\ \hline
Q2                       & 4.6   & 5.6     & 3.9 & \textbf{7.8} & 4.5    & 6.3    & 3.9  & \textbf{7.2} \\ \hline
Q3                       & 4.5   & 5.3     & 3.6 & \textbf{8.0} & 5.6    & 6.6    & 5.1  & \textbf{7.9} \\ \hline
Q4                       & 4.3   & 5.5     & 3.5 & \textbf{7.8} & 2.8    & 3.8    & 2.7  & \textbf{5.5} \\ \hline
Q5                       & 4.4   & 5.6     & 4.0 & \textbf{8.0} & 4.5    & 5.4    & 4.2  & \textbf{7.2} \\ \hline
Q6                       & 4.0   & 5.2     & 3.5 & \textbf{8.2} & 3.3    & 4.0    & 3.1  & \textbf{6.2} \\ \hline
Q7                       & 4.4   & 5.3     & 3.8 & \textbf{7.4} & 3.2    & 4.4    & 3.1  & \textbf{5.6} \\ \hline
Q8                       & 4.5   & 5.6     & 3.8 & \textbf{7.7} & 6.9    & 7.7    & 6.3  & \textbf{8.3} \\ \hline
Q9                       & 4.6   & 5.6     & 3.6 & \textbf{7.8} & 8.2    & 8.5    & 7.0  & \textbf{9.2} \\ \hline
Q10                      & 4.5   & 5.4     & 3.6 & \textbf{7.9} & 4.9    & 5.9    & 4.4  & \textbf{7.3} \\ \hline
Top-1                   & 0.0   & 0.1     & 0.0 & \textbf{0.9}  & \multicolumn{4}{c}{-} \\
\bottomrule
\end{tabular}
}
\label{tab:sea_e_eval}
\end{table}

%% file: table/human_eval_sea_s.tex



\begin{table}[hbtp]
\centering
\caption{Standardized content evaluation by humans and GPT.}
\large
\resizebox{\linewidth}{!}{
\begin{tabular}{c|c|c|c|c|c|c|c|c}
\toprule
\multirow{2}{*}{} & \multicolumn{4}{c|}{Human Evaluation} & \multicolumn{4}{c}{GPT Evaluation} \\ \cline{2-9} 
                        & \normalsize M-7B & \normalsize GPT-3.5 & \normalsize GPT-4   & \normalsize SEA-S        & \normalsize M-7B  & \normalsize GPT-3.5 & \normalsize GPT-4    & \normalsize SEA-S    \\ \hline

Q1   & 6.0  & 5.6  & 8.2  & \textbf{9.2} & 8.4  & 9.0  & 9.0  & \textbf{9.1} \\ \hline
Q2   & 6.7  & 7.5  & 7.6  & \textbf{8.3} & 8.9  & 8.8  & 8.7  & \textbf{9.3} \\ \hline
Q3   & 5.3  & 5.9  & 7.8  & \textbf{9.1} & 8.5  & 8.8  & 9.1  & \textbf{9.3} \\ \hline
Q4   & 7.3  & 3.7  & 7.3  & \textbf{9.0} & 8.7  & 7.1  & 8.0  & \textbf{8.9} \\ \hline
Q5   & 7.5  & 5.0  & 7.7  & \textbf{9.1} & 9.6  & 9.7  & 9.4  & \textbf{9.9} \\ \hline
Q6   & 6.7  & 6.8  & 6.9  & \textbf{8.1} & 9.3  & 9.6  & 9.4  & \textbf{9.9} \\ \hline
Q7   & 5.1  & 5.2  & 8.1  & \textbf{8.8} & 8.5  & 9.0  & 9.0  & \textbf{9.2} \\ \hline
\end{tabular}
}
\label{tab:sea_s_eval}
\end{table}

%% file: sec-conclusion.tex
\section{Conclusion}
In this paper, we present SEA, a novel framework for automated paper reviewing.
Specifically, 
we propose a new paradigm for constructing a standardized 
review
dataset.
Based on this dataset,
we fine-tune an LLM to generate high-quality reviews.
Moreover, 
we propose a new evaluation metric to measure the consistency between papers and generated reviews. 
Comprehensive experimental results demonstrate that the SEA framework can generate feedback that aligns with human reviews.
In summary, we emphasize that  
the initial motivation for the field of automated paper review stems from the time-consuming and labor-intensive nature of traditional peer reviewing. 
Automated peer reviewing can provide timely feedback, enhancing research quality and accelerate the progress of scientific development. 
Therefore, we anticipate that the SEA framework will help researchers improve the quality of their work and shed light on the field of automated scientific reviewing.
\clearpage

%% file: sec-limitations.tex
\section*{Limitations}
Despite these notable achievements, it is crucial to acknowledge the limitations of SEA,
particularly its limited expansion into various academic disciplines and insufficient alignment with human standards.
Here we elaborate on some of these constraints, along with intriguing future  explorations.

\paragraph{Domain Expansion.}
Although the SEA framework has been successful in automating paper review generation within the machine learning field, it has not yet been expanded to other academic disciplines, such as physics and mathematics.
As a universal automated paper review framework,
SEA is able to generalize across any field. 
Thus, it would be exhilarating to investigate whether SEA can yield high-quality review feedback when applied to other academic disciplines.

\paragraph{Enhanced Consistency-Guided Training.}
Although optimizing the output of SEA-E by calculating mismatch scores between review and the original paper 
can generate review that are more consistent in content, 
we did not enhance SEA-E using natural language guidance based on scores during the training phase. 
To improve SEA-E in following instructions during the self-correction phase, 
we plan to collect relevant natural language guided self-correction dataset. 
By training on this dataset, we will further enhance SEA-E in content preference, 
enabling it to generate review feedback that aligns more accurately with the original paper.

\paragraph{Rebuttal Exploration.}

In the academic peer review process, the rebuttal stage is a critical component. 
During this stage, 
authors have the opportunity to correct potential misunderstandings by reviewers, clarify specific parts of their paper, or provide additional data and information to enhance the support for their research findings. Therefore, in our future research, we will explore methods to assist authors in making effective rebuttals.

%% file: sec-ethics.tex
\section*{Ethical Considerations}
This paper proposes an automated paper reviewing framework that utilizes advanced long-context LLMs and supervised fine-tuning to align with human reviews and generate comprehensive reviews.
This assists authors in improving the quality of their papers.
As we explore the extensive potential of automated paper reviewing, 
it is essential to consider potential consequences associated with this technology.
A significant concern is the misuse of the model.
In the formal review processes of academic conferences,
authors may receive reviews generated by the model without their knowledge. 
This situation could not only impact the fairness and transparency of the review process but also raise issues of trust and authenticity.
To mitigate these risks, 
we will incorporate specific clauses in our usage license that strictly prohibit any misuse of the system, thereby ensuring it serves as a beneficial tool in academia.


%% file: sec-appendix.tex
\clearpage

\section{More Detailed Description of the Framework SEA }
\label{sec:appendix}

\subsection{SEA-S}
\label{app:sea-s}
We further analyse the performance of SEA-S, the open-source model Mistral-7B, and the closed-source models GPT-3.5 and GPT-4 in standardised review experiments.
\paragraph{Instruction.} 
In Table~\ref{tab:instruction_sea-s},
we demonstrate our instructions for generating standardized review based on multiple reviews for each paper.
We specify in the instruction that the model should integrate multiple reviews into three parts: 
textual descriptions, quantitative scores, and review results.
The textual descriptions include ``Summary'', ``Strengths'', ``Weaknesses'', and ``Questions'', while the quantitative scores cover ``Soundness'', ``Presentation'', ``Contribution'', and ``Rating''. 
These elements are formatted in alignment with the original review template.
Additionally, we incorporate the Area Chair's (AC) decision into the generated content, and instruct the model to generate corresponding acceptance or rejection reasons.

\begin{figure*}[h]
    \centering
\includegraphics[width=1.0\linewidth]{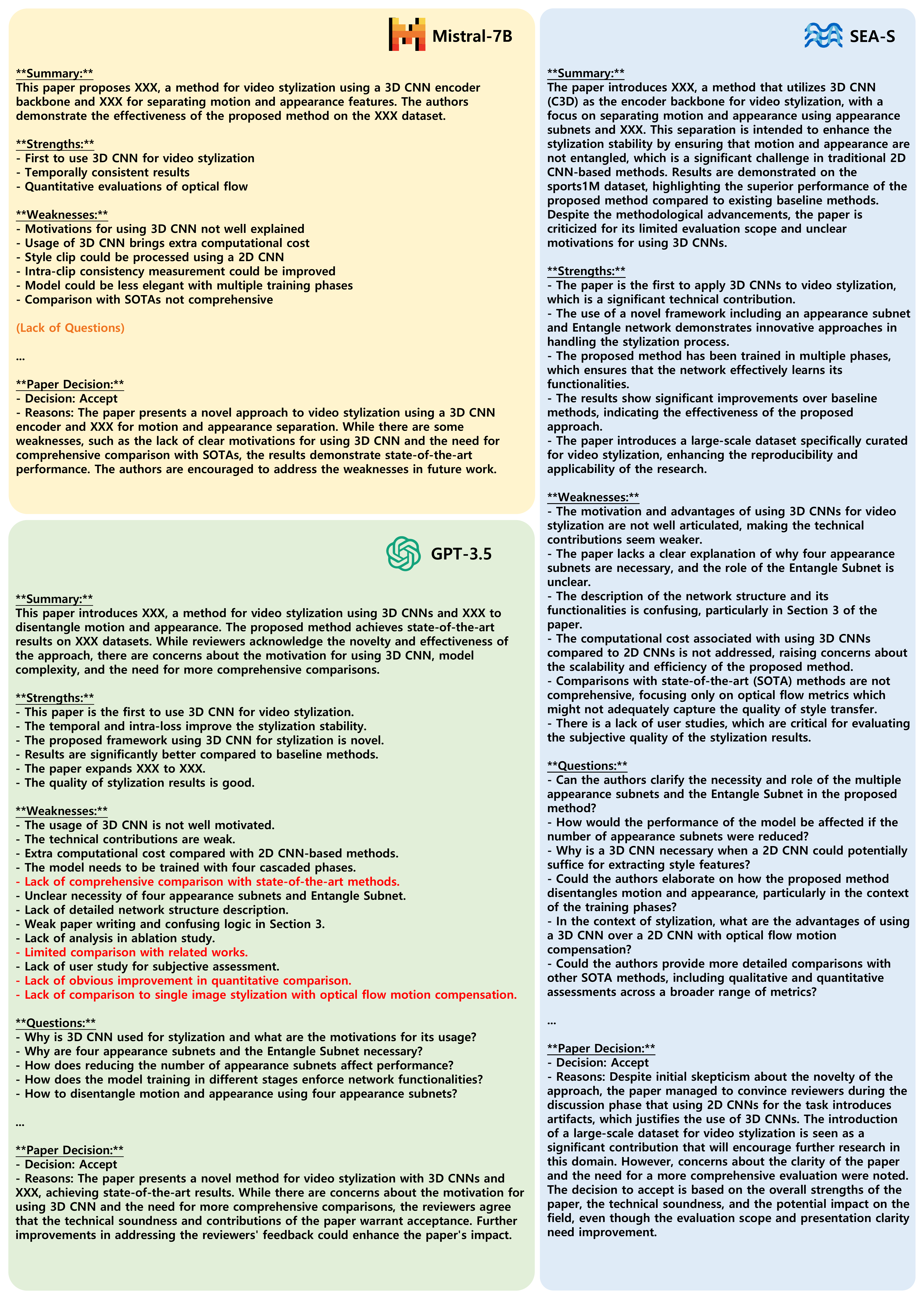}
    \caption{Examples of standardization for Mistral-7B, GPT-3.5 and SEA-S.}
    \label{fig:sea-s_exmaple}
\end{figure*}

\paragraph{Standardization Examples.}
Figure~\ref{fig:sea-s_exmaple} shows standardization examples from Mistral-7B, GPT-3.5, and SEA-S,
which incorporate multiple reviews for the same paper.
We can observe from the figure that the output of SEA-S is both rich and concise without redundant information.
In contrast,
the output from Mistral-7B not only lacks complete format but also has sparse content, with the missing parts highlighted in orange in the figure.
As for the review generated by GPT-3.5, 
a significant portion consists merely of straightforwardly extracting original review content, failing to eliminate redundant information as instructed,
such as the overuse of the phrase ``Lack of'', which is indicated in red to show the excessive repetition.

\subsection{SEA-E}
\label{app:sea-e}
\begin{figure*}[h]
    \centering
\includegraphics[width=1.0\linewidth]{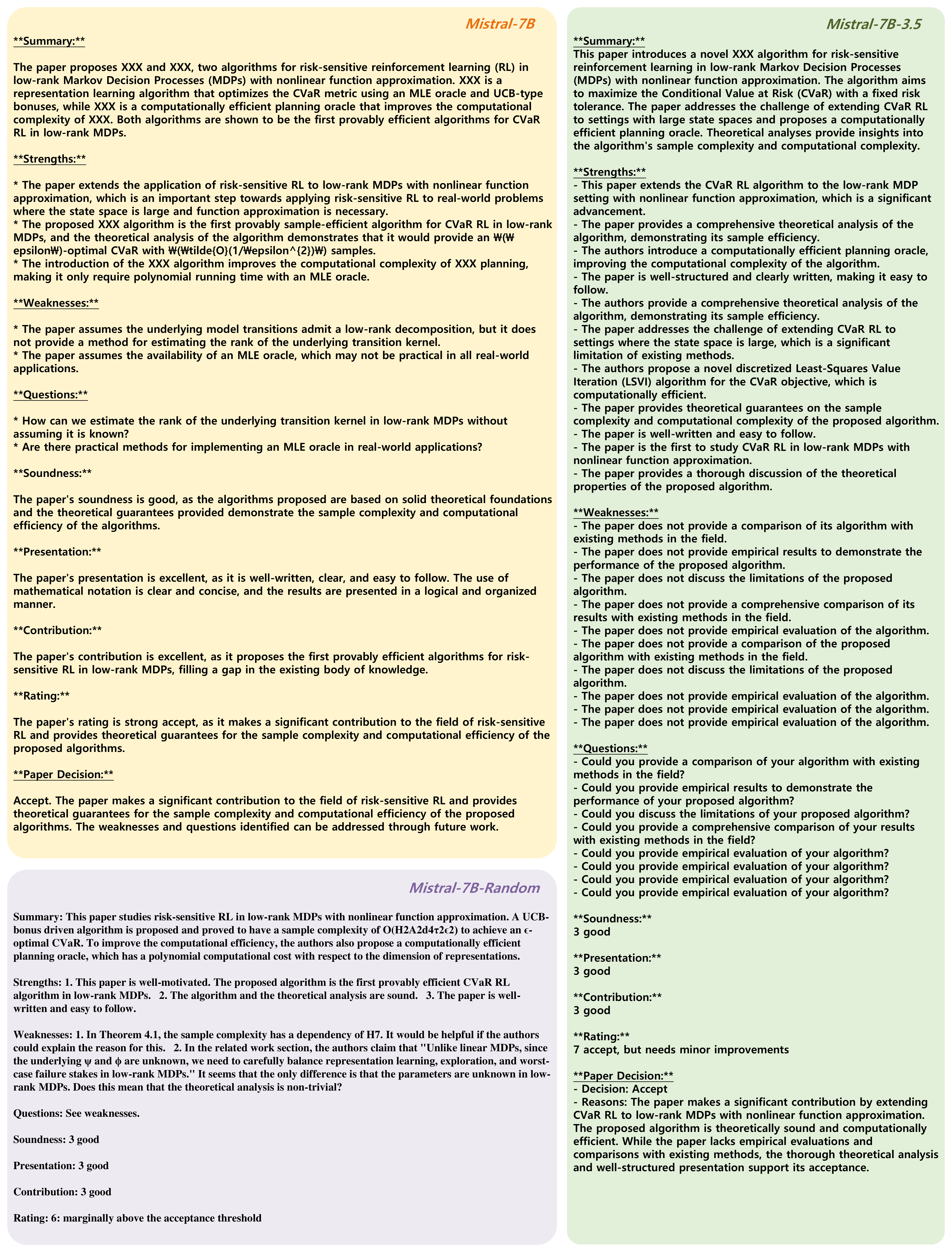}
    \caption{Examples of generated reviews for Mistral-7B, Mistral-7B-Random and Mistral-7B-GPT-3.5.}
    \label{fig:sea-e_exmaple1}
\end{figure*}

\begin{figure*}[h]
    \centering
\includegraphics[width=1.0\linewidth]{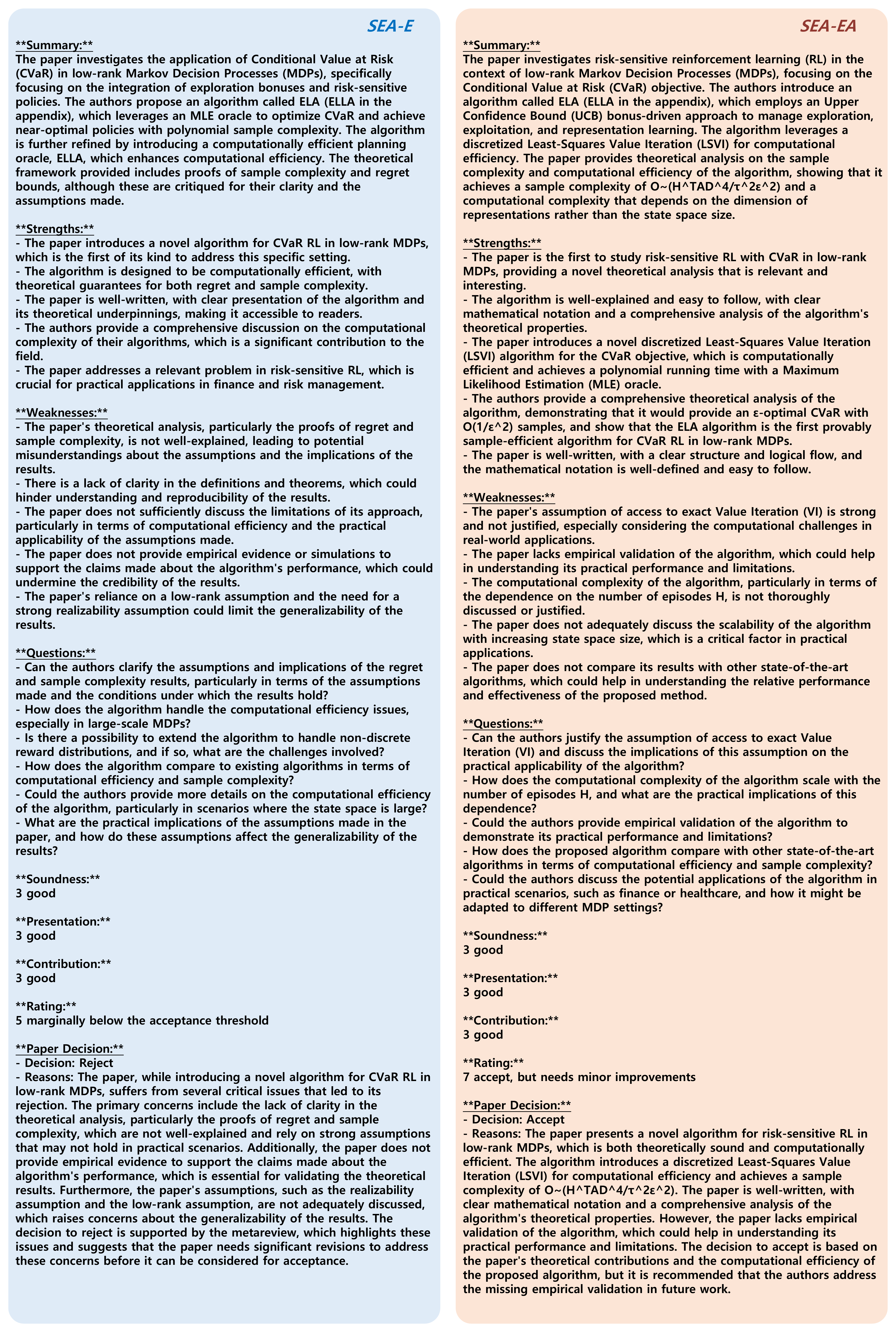}
    \caption{Examples of generated reviews for SEA-E and SEA-EA.}
    \label{fig:sea-e_exmaple2}
\end{figure*}

In Table~\ref{tab:instruction_sea-e}, we present the instruction designed to generate reviews that conform to the specified format based on the content of the paper.
In Figures~\ref{fig:sea-e_exmaple1} and~\ref{fig:sea-e_exmaple2}, 
we display the reviews generated by different models for a particular paper, including Mistral-7B (M-7B), Mistral-7B-Random (M-7B-R), Mistral-7B-GPT-3.5 (M-7B-3.5), SEA-E, and SEA-EA. We can observe the following points:
(1)	Mistral-7B raises broad and general issues, tending to please humans. 
In the ``Strengths'' part, it splits the complexity issue into two points, 
which is not concise, 
and the content of the ``Weaknesses'' part does not match the paper decision.
(2)	Mistral-7B-Random visibly generates shorter texts with reduced detail.
(3)	Mistral-7B-GPT-3.5 generates duplicates due to insufficient standardization of the instruction dataset at the SFT stage, resulting in lower-quality reviews.
(4)	SEA-E and SEA-EA generate clearer viewpoints and ensure extensive coverage of content.
(5)	SEA-EA focuses more on the details within the paper.
These comparisons demonstrate the superiority of SEA-E and SEA-EA in generating reviews.

\subsection{SEA-A}
\label{app:sea-a}
To demonstrate the effectiveness of the regression model SEA-A, 
we randomly select a review for each paper from each dataset to form a paper-review pair.
Then,
we use SEA-A to calculate mismatch score, 
which is displayed in Table~\ref{tab:sea-a-random}.
Since SEA-A is trained with a majority of low-scoring samples, the values of the mismatch scores 
are not substantial.
To enhance the intuitiveness of the main text,
we present the results as Figure~\ref{fig:mms}, 
and here Table~\ref{tab:sea_a} demonstrates the specific values instead.
By comparing Table~\ref{tab:sea-a-random} with Table~\ref{tab:sea_a}, each element of the former is larger than the corresponding item of the latter. 
Therefore, our regression model has the ability to discern the consistency between different reviewers and papers.

\input{table/sea-a_random}
\input{table/sea-a_result}

\input{table/r2_exp}

\begin{figure*}[h]
    \centering
\includegraphics[width=1.0\linewidth]{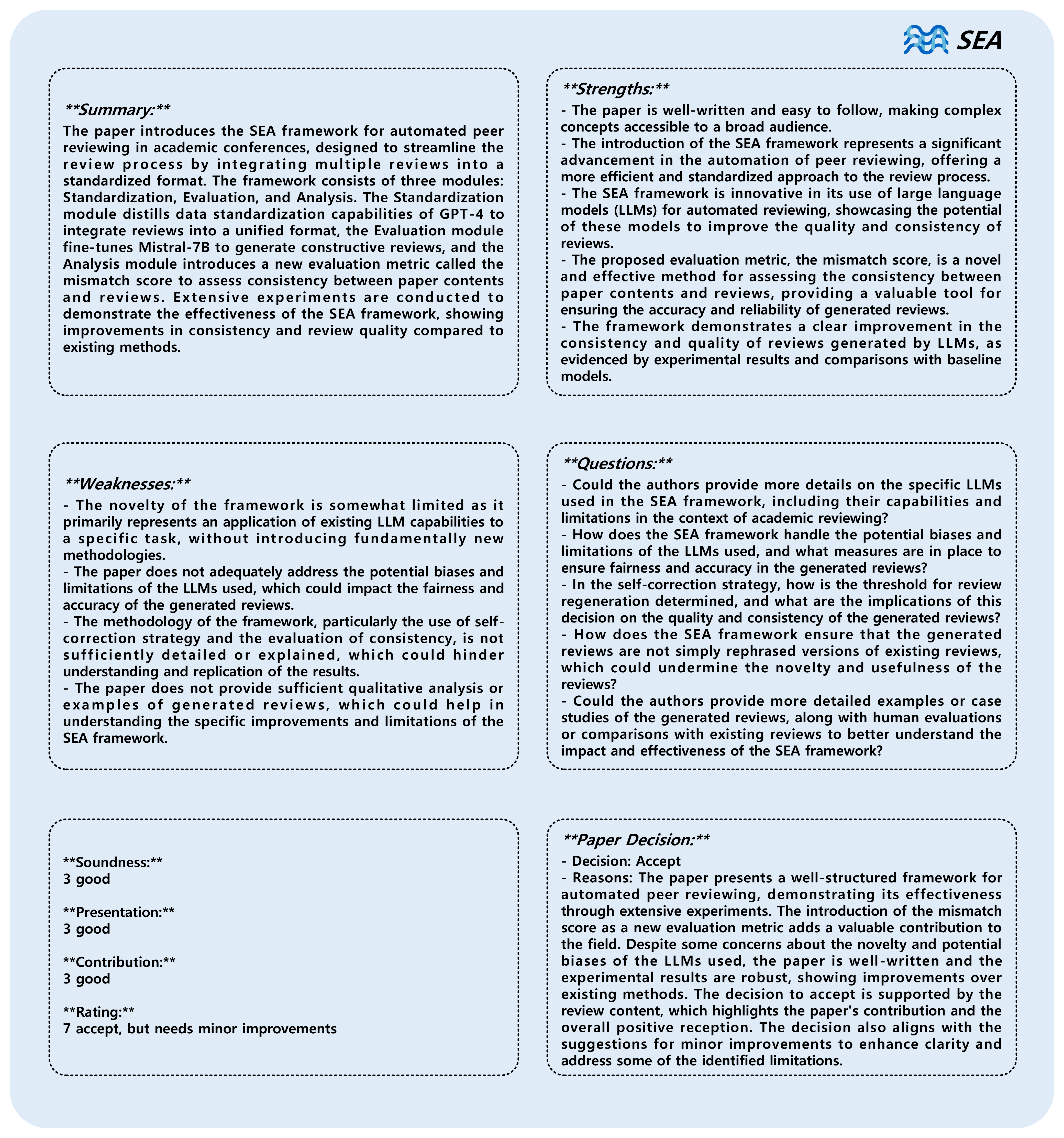}
    \caption{The review generated by applying the SEA framework to this paper.}
    \label{fig:sea-self}
\end{figure*}

\emph{
Interestingly, we apply the SEA framework to this paper and compare its generated review with the official feedback we receive. 
We find that the SEA framework aligns with some aspects of the actual review in terms of ``Strengths,'' ``Weaknesses,'' and ``Questions.''
In addition, SEA provides further constructive suggestions for improvement, 
demonstrating its ability to generate comprehensive and high-quality review comments. 
The specific generated reviews are shown in Figure~\ref{fig:sea-self}.}


\section{Compare with REVIEWER2}

\label{app:reviewer2}

To further validate the effectiveness of our framework SEA, 
we compare its performance with the open-source model of REVIEWER2~\cite{reviewer2}. 
Given that using two LLMs for inference process of REVIEWER2 is more time-consuming,
we sample a smaller test set which is a subset of the test set used in this paper.
Specifically, we randomly choose 100 samples from each dataset (or use all samples if the dataset contains fewer than 100). 
When inferring the model of REVIEWER2\footnote{https://github.com/ZhaolinGao/Reviewer2}, 
we follow the settings described in the original paper. 
Table~\ref{tab:r2_result} lists the results for REVIEWER2 (abbreviated as R2), other baseline models, and our proposed framework.
The results show that both SEA-EA and SEA-E exhibit excellent performance.
In contrast, the results for REVIEWER2 are not ideal in the ROUGE metric and are unstable in the BERTScore metric.
This is because REVIEWER2 often generates contents that are relatively short and lack valuable information.
In contrast, our methods which fine-tune on a high-quality instruction dataset can generate more comprehensive reviews, 
demonstrating the superiority of our framework.

\clearpage
\section{Human and GPT evaluation}
\label{app:questionnair}

In order to evaluate the review quality generated by SEA-E and the standardized content quality generated by SEA-S, we design a questionnaire consisting of 10 questions and 7 questions, respectively. For the review quality, we add an additional question during the human evaluation phase: selecting the best review from those generated by different models. 
The specific questionnaire contents are as follows:

\paragraph{SEA-E.}
\begin{enumerate}[leftmargin=*, itemsep=0pt, parsep=0pt, topsep=0pt, partopsep=0pt]
    \item Did the reviewer discuss the importance of the research question?
    \item Did the reviewer discuss the originality of the paper?
    \item Did the reviewer clearly identify the strengths and weaknesses of the method (study design, data collection and data analysis)?
    \item Did the reviewer make specific useful comments on the writing, organization, tables and figures of the manuscript?
    \item Were the reviewer's comments constructive?
    \item Did the reviewer supply appropriate evidence using examples from the paper to substantiate their comments?
    \item Did the reviewer comment on the author's interpretation of the results?
    \item Did the reviewer maintain objectivity and avoid bias in the review?
    \item Was the language used by the reviewer professional and appropriate?
    \item How would you rate the quality of this review overall?
    \item Please select what you consider to be the best review. (The question only in the human evaluation)
\end{enumerate}

\paragraph{SEA-S.}

\begin{enumerate}[leftmargin=*, itemsep=0pt, parsep=0pt, topsep=0pt, partopsep=0pt]
    \item Did this review effectively consolidate similar points from the original reviews?
    \item Did the review maintain objectivity and avoid bias?
    \item Did this review contain a lot of useful information?
    \item Did this review contain a lot of repetitive content?
    \item Is there any contradiction in this review?
    \item Was the language used by the review professional and appropriate?
    \item How would you rate the quality of this review overall?
\end{enumerate}

\begin{table*}[ht]
\centering
  \caption{Instruction for generating standardized review based on multiple reviews for each paper.}
  {
  \begin{tabular}{p{15cm}}
    \hline
    \hline
\textbf{INSTRUCTION:~}
\\
\quad  As an experienced academic paper reviewer, you are presented with different review contents for the same paper. 
Please analyze these contents carefully and consolidate them into a single review. The review should be organized into nine sections: Summary, Strengths, Weaknesses, Questions, Soundness, Presentation, Contribution, Rating and Paper Decision. 
Below is a description of each section:\\
1. Summary: Combine the `Summary' sections from all reviews into a cohesive summary, aiming for a length of about 100-150 words.\\
2. Strengths/Weaknesses/Questions: Combine the Strengths/Weaknesses/Questions sections from all reviews into a unified, cohesive bullet-point list that avoids redundancy while preserving the specific details and depth of each point.\\
3. Soundness/Presentation/Contribution: Aggregate the Contribution/Soundness/Presentation score from each review to determine a suitable overall score (the score must be an **integer**), then, match this integer score to the corresponding criterion from the list below and provide the result. For example, if the score is 3, the result should be `3 good'. The possible scores and their criteria are: \\
\qquad 1 poor \textbackslash n 
\qquad 2 fair \textbackslash n 
\qquad 3 good \textbackslash n 
\qquad 4 excellent \\
4. Rating: Aggregate the `Rating' from each review to determine a suitable overall Rating (the Rating must be an **integer**), then, match this integer Rating to the corresponding criterion from the list below and provide the result. For example, if the Rating is 1, the result should be `1 strong reject'. The possible Ratings and their criteria are: \\
\qquad 1 strong reject \\
\qquad 2 reject, significant issues present \\
\qquad 3 reject, not good enough \\
\qquad 4 possibly reject, but has redeeming facets \\
\qquad 5 marginally below the acceptance threshold \\
\qquad 6 marginally above the acceptance threshold \\
\qquad 7 accept, but needs minor improvements \\
\qquad 8 accept, good paper \\
\qquad 9 strong accept, excellent work\\
\qquad 10 strong accept, should be highlighted at the conference   \\
5. Paper Decision: It must include the Decision itself~(Accept or Reject) and the reasons for this decision which is based on Meta-review, the criteria of originality, methodological soundness, significance of results, and clarity and logic of presentation, etc. Please ensure your Decision (Accept/Reject) matches the value of the `Decision' key in the JSON, if present. \\\\
Here is the template for a review format, you must follow this format to output your review result:\\ 
**Summary:** \quad \textbackslash n 
<Summary content> \textbackslash n \\\\
**Strengths:**
\quad \textbackslash n <Strengths result> \textbackslash n
\\
**Weaknesses:**
\quad \textbackslash n <Weaknesses result> \textbackslash n
\\
**Questions:** 
\quad \textbackslash n <Questions result> \textbackslash n
\\\\
**Soundness:** \quad \textbackslash n <Soundness result> \textbackslash n \\
**Presentation:** \quad \textbackslash n <Presentation result> \textbackslash n \\
**Contribution:** \quad \textbackslash n <Contribution result> \textbackslash n \\
**Rating:** \quad \textbackslash n <Rating result> \textbackslash n \\\\
**Paper Decision:**\\- Decision: Accept/Reject\\- Reasons: reasons content\\\\

       \hline
    \hline
  \end{tabular}}
  \label{tab:instruction_sea-s}
\end{table*}

\begin{table*}[ht]
\centering
  \caption{Instructions for generating review comments based on the content of the paper.}
  {
  \begin{tabular}{p{15cm}}
    \hline
    \hline
\textbf{INSTRUCTION:~}
\\
\quad  You are a highly experienced, conscientious, and fair academic reviewer. 
Please help me review this paper. 
The review should be organized into nine sections: \\
1. Summary: A summary of the paper in 100-150 words. \\
2. Strengths/Weaknesses/Questions: The Strengths/Weaknesses/Questions of paper, which should be listed in bullet points, with each point supported by specific examples from the article where possible. \\
3. Soundness/Contribution/Presentation: Rate the paper's Soundness/Contribution/Presentation, and match this score to the corresponding criterion from the list below and provide the result. The possible scores and their criteria are: \\
\qquad 1 poor \\
\qquad 2 fair \\
\qquad 3 good \\
\qquad 4 excellent \\
4. Rating: Give this paper an appropriate rating, match this rating to the corresponding criterion from the list below and provide the result. The possible Ratings and their criteria are: \\
\qquad 1 strong reject \\
\qquad 2 reject, significant issues present \\
\qquad 3 reject, not good enough \\
\qquad 4 possibly reject, but has redeeming facets \\
\qquad 5 marginally below the acceptance threshold \\
\qquad 6 marginally above the acceptance threshold \\
\qquad 7 accept, but needs minor improvements \\
\qquad 8 accept, good paper \\
\qquad 9 strong accept, excellent work\\
\qquad 10 strong accept, should be highlighted at the conference   \\
5. Paper Decision: It must include the Decision itself
(Accept or Reject) and the reasons for this decision which is based on the criteria of originality, methodological soundness, significance of results, and clarity and logic of presentation.\\\\
Here is the template for a review format, you must follow this format to output your review result:\\ 
**Summary:** \quad \textbackslash n 
<Summary content> \textbackslash n \\\\
**Strengths:**
\quad \textbackslash n <Strengths result> \textbackslash n
\\
**Weaknesses:**
\quad \textbackslash n <Weaknesses result> \textbackslash n
\\
**Questions:** 
\quad \textbackslash n <Questions result> \textbackslash n
\\\\
**Soundness:** \quad \textbackslash n <Soundness result> \textbackslash n \\
**Presentation:** \quad \textbackslash n <Presentation result> \textbackslash n \\
**Contribution:** \quad \textbackslash n <Contribution result> \textbackslash n \\
**Rating:** \quad \textbackslash n <Rating result> \textbackslash n \\\\
**Paper Decision:**\\- Decision: Accept/Reject\\- Reasons: reasons content\\\\
Please ensure your feedback is objective and constructive. The paper is as follows: <paper content>

\\
       \hline
    \hline
  \end{tabular}}
  \label{tab:instruction_sea-e}
\end{table*}

%% file: table/sea-a_random.tex
\begin{table}[h]
  \centering
  \caption{Performance of mismatch scores in random pairs of papers and reviews.}
    \resizebox{\linewidth}{!}{
    \begin{tabular}{l|ccccc}
    \toprule
    Datasets & M-7B  & M-7B-3.5 & M-7B-R & SEA-E & SEA-EA \\
    \midrule
    CONLL-16 & 1.1974  & 1.0118  & 0.5904  & 0.5832  & 0.5057  \\
    ACL-17 & 1.0146  & 0.6658  & 0.5784  & 0.4855  & 0.5006  \\
    COLING-20 & 0.9731  & 0.5553  & 0.4699  & 0.4733  & 0.4420  \\
    ARR-22 & 0.8285  & 0.5656  & 0.5262  & 0.4452  & 0.4043  \\
    \midrule
    NeurIPS-16-22 & 0.9640  & 0.8343  & 0.6974  & 0.5792  & 0.5536  \\
    ICLR-17-23 & 0.9850  & 0.6169  & 0.6551  & 0.4755  & 0.4474  \\
    NeurIPS-23 & 0.9451  & 0.7252  & 0.7022  & 0.5964  & 0.5513  \\
    ICLR-24 & 0.9348  & 0.6037  & 0.5935  & 0.4256  & 0.3999  \\
    \bottomrule
    \end{tabular}}
  \label{tab:sea-a-random}%
\end{table}%

%% file: table/sea-a_result.tex
\begin{table}[h]
  \centering
  \caption{Performance of mismatch score in pairs of papers and corresponding reviews.}
    \resizebox{\linewidth}{!}{
    \begin{tabular}{cl|ccccc}
    \toprule
    \multicolumn{2}{c|}{Datasets} & M-7B  & M-7B-R & M-7B-3.5 & SEA-E & SEA-EA \\
    \midrule
    \multirow{4}[2]{*}{\begin{sideways}\footnotesize{Cross-domain}\end{sideways}} & CONLL-16 & 1.0503  & 0.8416  & 0.5665  & 0.5595  & 0.3926 \\
          & ACL-17 & 0.9309  & 0.6608  & 0.5257  & 0.4529  & 0.4359 \\
          & COLING-20 & 0.8642  & 0.5235  & 0.4446  & 0.3931  & 0.3888 \\
          & ARR-22 & 0.7095  & 0.5136  & 0.4964  & 0.3953  & 0.3926 \\
    \midrule
    \multirow{4}[2]{*}{\begin{sideways}\footnotesize{In-domain}\end{sideways}} & NeurIPS-16-22 & 0.8409  & 0.7282  & 0.6271  & 0.5098  & 0.4733 \\
          & ICLR-17-23 & 0.8630  & 0.5759  & 0.5924  & 0.4358  & 0.4227 \\
          & NeurIPS-23 & 0.7638  & 0.6511  & 0.6388  & 0.5109  & 0.4541 \\
          & ICLR-24 & 0.7746  & 0.5203  & 0.5396  & 0.4063  & 0.3788 \\
    \bottomrule
    \end{tabular}%
    }
  \label{tab:sea_a}%
\end{table}

%% file: table/r2_exp.tex
\begin{table*}[t]
\caption{The overall performance (\%) on the smaller test set.
}
    \captionsetup{justification=centering}
    \begin{minipage}{.5\linewidth}
      \centering
        \resizebox{\linewidth}{!}{

    \begin{tabular}{l|c|ccc|ccc|c}
    \toprule
    \multicolumn{1}{c|}{\multirow{2}[2]{*}{Method}} & \multicolumn{1}{c}{\multirow{2}[2]{*}{BLEU}} & \multicolumn{3}{c|}{ROUGE (Recall)} & \multicolumn{3}{c|}{ROUGE (F1-score)} & \multirow{2}[2]{*}{BERTScore} \\
          & \multicolumn{1}{c}{} & R-1   & R-2   & R-L   & R-1   & R-2   & R-L   &  \\
    \midrule
    \multicolumn{9}{c}{\textit{CONLL-16}} \\
    R2    & 15.21  & 17.15  & 4.27  & 8.63  & 25.24  & 6.40  & 12.67  & \textbf{83.00} \\
    M-7B  & 18.92  & 20.81  & 4.81  & 10.30  & 28.66  & 6.81  & 14.18  & 82.49  \\
    M-7B-R & 18.16  & 21.96  & 5.17  & 10.62  & 29.56  & 7.18  & 14.31  & 82.57  \\
    M-7B-3.5 & 19.70  & 26.51  & 5.58  & 13.96  & 30.19  & 6.45  & 15.37  & 82.01  \\
    SEA-E & \underline{29.07}  & \underline{34.91}  & \underline{7.79}  & \underline{15.29}  & \underline{38.64}  & \underline{8.67}  & \underline{16.73}  & 82.91  \\
    SEA-EA & \textbf{31.01} & \textbf{36.96} & \textbf{8.91} & \textbf{16.34} & \textbf{40.49} & \textbf{9.68} & \textbf{17.57} & \underline{82.94}  \\
    \midrule
    \multicolumn{9}{c}{\textit{ACL-17}} \\
    R2    & 14.20  & 17.66  & 4.42  & 8.89  & 23.86  & 6.25  & 12.07  & 82.26  \\
    M-7B  & 18.37  & 21.32  & 4.92  & 10.50  & 27.39  & 6.47  & 13.38  & 82.56  \\
    M-7B-R & 17.93  & 22.14  & 5.15  & 10.84  & 27.50  & 6.72  & 13.34  & 82.47  \\
    M-7B-3.5 & 16.23  & 27.35  & 6.13  & 14.68  & 25.87  & 5.99  & 13.15  & 82.23  \\
    SEA-E & \underline{24.86}  & \underline{33.02}  & \underline{7.51}  & \underline{14.97}  & \underline{34.97}  & \underline{8.15}  & \underline{15.38}  & \underline{82.87}  \\
    SEA-EA & \textbf{27.02} & \textbf{35.66} & \textbf{8.61} & \textbf{15.85} & \textbf{37.48} & \textbf{9.16} & \textbf{16.11} & \textbf{83.05} \\
    \midrule
    \multicolumn{9}{c}{\textit{COLING-20}} \\
    R2    & 18.08  & 23.71  & 5.49  & 12.14  & 28.57  & 6.75  & 14.60  & 82.04  \\
    M-7B  & 21.97  & 29.11  & 6.42  & 14.80  & 31.91  & 7.01  & 15.83  & 82.76  \\
    M-7B-R & 19.49  & 29.21  & 6.69  & 15.20  & 30.23  & 6.80  & 15.25  & 82.27  \\
    M-7B-3.5 & 18.13  & 34.03  & 7.56  & 18.43  & 28.49  & 6.10  & 14.77  & 82.12  \\
    SEA-E & \underline{22.93}  & \underline{40.62}  & \underline{9.23}  & \underline{20.05}  & \underline{34.37}  & \underline{7.65}  & \underline{16.15}  & \underline{82.84}  \\
    SEA-EA & \textbf{24.85} & \textbf{42.97} & \textbf{10.57} & \textbf{20.89} & \textbf{36.67} & \textbf{8.76} & \textbf{16.96} & \textbf{83.09} \\
    \midrule
    \multicolumn{9}{c}{\textit{ARR-22}} \\
    R2    & 17.87  & 22.62  & 6.20  & 11.83  & 28.62  & 8.13  & 15.03  & 79.29  \\
    M-7B  & 23.74  & 28.81  & 7.99  & 14.56  & 34.31  & \underline{9.71}  & 17.26  & 83.41  \\
    M-7B-R & 21.77  & 28.49  & 7.66  & 14.86  & 32.60  & 8.98  & 16.84  & 82.72  \\
    M-7B-3.5 & 18.55  & 34.27  & 8.55  & 18.47  & 29.47  & 7.64  & 15.20  & 82.65  \\
    SEA-E & \underline{25.27}  & \underline{40.40}  & \underline{10.24}  & \underline{19.40}  & \underline{37.68}  & 9.70  & \underline{17.50}  & \underline{83.46}  \\
    SEA-EA & \textbf{27.16} & \textbf{43.02} & \textbf{11.93} & \textbf{20.27} & \textbf{39.94} & \textbf{11.21} & \textbf{18.30 } & \textbf{83.66} \\
    \bottomrule
    \end{tabular}%

}
    
    \end{minipage}%
    \hfill
    \begin{minipage}{.49\linewidth}
      \centering
        \resizebox{\linewidth}{!}{

    \begin{tabular}{l|c|ccc|ccc|c}
    \toprule
    \multicolumn{1}{c|}{\multirow{2}[2]{*}{Method}} & \multirow{2}[2]{*}{BLEU} & \multicolumn{3}{c|}{ROUGE (Recall)} & \multicolumn{3}{c|}{ROUGE (F1-score)} & \multirow{2}[2]{*}{BERTScore} \\
          &       & R-1   & R-2   & R-L   & R-1   & R-2   & R-L   &  \\
    \midrule
    \multicolumn{9}{c}{\textit{NeurIPS-16-22}} \\
    R2    & 10.41  & 11.00  & 3.94  & 5.95  & 18.23  & 6.64  & 9.88  & \textbf{83.30} \\
    M-7B  & 14.94  & 14.85  & 5.08  & 7.44  & 23.47  & 8.05  & 11.73  & 82.91  \\
    M-7B-R & 12.86  & 14.14  & 4.78  & 7.46  & 21.65  & 7.52  & 11.22  & 82.56  \\
    M-7B-3.5 & 16.48  & 21.43  & 6.33  & \underline{11.34}  & 26.36  & 8.12  & 13.49  & 82.34  \\
    SEA-E & \underline{25.03}  & \underline{24.82}  & \underline{7.38}  & 10.98  & \underline{34.59}  & \underline{10.41}  & \underline{15.30}  & 83.14  \\
    SEA-EA & \textbf{27.16} & \textbf{27.43} & \textbf{8.60} & \textbf{11.98} & \textbf{37.32} & \textbf{11.77} & \textbf{16.26} & \underline{83.28}  \\
    \midrule
    \multicolumn{9}{c}{\textit{ICLR-17-23 }} \\
    R2    & 9.19  & 9.25  & 3.51  & 5.06  & 15.94  & 6.09  & 8.78  & 83.39  \\
    M-7B  & 13.53  & 12.93  & 4.54  & 6.46  & 21.50  & 7.59  & 10.78  & 83.22  \\
    M-7B-R & 12.83  & 12.95  & 4.32  & 6.60  & 21.02  & 7.13  & 10.74  & 82.66  \\
    M-7B-3.5 & 16.22  & 19.10  & 5.75  & \underline{10.14}  & 25.71  & 7.98  & 13.16  & 82.73  \\
    SEA-E & \underline{23.21}  & \underline{22.17}  & \underline{6.88}  & 9.89  & \underline{32.31}  & \underline{10.14}  & \underline{14.47}  & \underline{83.48}  \\
    SEA-EA & \textbf{25.29} & \textbf{24.70} & \textbf{7.95} & \textbf{10.75} & \textbf{35.17} & \textbf{11.45} & \textbf{15.37} & \textbf{83.62} \\
    \midrule
    \multicolumn{9}{c}
    {\textit{NeurIPS-23}} \\
    R2    & 7.84  & 8.29  & 3.33  & 4.63  & 14.68  & 5.91  & 8.23  & 83.19  \\
    M-7B  & 12.84  & 12.35  & 5.13  & 6.36  & 21.17  & 8.81  & 10.92  & 84.00  \\
    M-7B-R & 12.34  & 12.18  & 4.93  & 6.27  & 20.57  & 8.36  & 10.65  & 83.68  \\
    M-7B-3.5 & 16.33  & 17.39  & 6.33  & 8.89  & 26.29  & 9.73  & 13.29  & 83.28  \\
    SEA-E & \underline{21.86}  & \underline{20.81}  & \underline{7.46}  & \underline{9.38}  & \underline{31.98}  & \underline{11.49}  & \underline{14.45}  & \underline{84.13}  \\
    SEA-EA & \textbf{23.78} & \textbf{22.91} & \textbf{8.60} & \textbf{10.12} & \textbf{34.59} & \textbf{13.02} & \textbf{15.31} & \textbf{84.31} \\
    \midrule
    \multicolumn{9}{c}{\textit{ICLR-24}} \\
    R2    & 8.91  & 9.26  & 3.61  & 5.06  & 16.16  & 6.34  & 8.88  & 83.30  \\
    M-7B  & 13.25  & 12.74  & 4.90  & 6.37  & 21.50  & 8.30  & 10.78  & 83.98  \\
    M-7B-R & 13.47  & 13.69  & 5.23  & 6.96  & 22.16  & 8.57  & 11.28  & 83.89  \\
    M-7B-3.5 & 16.88  & 20.21  & 6.68  & \underline{10.56}  & 27.32  & 9.34  & 13.79  & 83.44  \\
    SEA-E & \underline{23.06}  & \underline{22.58}  & \underline{7.62}  & 9.84  & \underline{33.57}  & \underline{11.38}  & \underline{14.68}  & \underline{84.05}  \\
    SEA-EA & \textbf{25.44} & \textbf{25.19} & \textbf{8.81} & \textbf{10.70} & \textbf{36.62} & \textbf{12.88} & \textbf{15.61} & \textbf{84.23} \\
    \bottomrule
    \end{tabular}%

        }
    \end{minipage}
        
    \label{tab:r2_result}
\end{table*}